\pdfoutput=1

\documentclass[11pt]{article}

\usepackage[preprint]{acl}

\usepackage{times}
\usepackage{latexsym}
\usepackage{placeins}
\usepackage{amsfonts}
\usepackage{colortbl}
\usepackage{multirow}
\usepackage{booktabs}
\usepackage{arydshln}
\usepackage{caption}
\usepackage{hhline}
\usepackage{amssymb}
\usepackage{makecell}
\usepackage{xcolor}
\usepackage{subcaption}
\usepackage{listings}
\usepackage{comment}
\usepackage{float}
\usepackage{pifont}
\newcommand{\fstar}{\ding{72}} %
\newcommand{\estar}{\ding{73}} %

\usepackage[T1]{fontenc}
\usepackage[utf8]{inputenc}

\usepackage{microtype}

\usepackage{inconsolata}

\usepackage{graphicx}

\title{Rubric Spans are Label Representations: Joint LLM Encoding for Short Answer Scoring}

\author{
  Zhifan Sun\textsuperscript{1} \quad Sebastian Gombert\textsuperscript{1} \quad Fabian Zehner\textsuperscript{1,2} \\
  \textbf{Leon Camus\textsuperscript{1} \quad Longwei Cong\textsuperscript{1} \quad Hendrik Drachsler\textsuperscript{1,3}} \\
  \textsuperscript{1}DIPF \textbar{} Leibniz Institute for Research and Information in Education \\
  \textsuperscript{2}Centre for International Student Assessment (ZIB) \\
  \textsuperscript{3}Computer Science Department \& Studiumdigitale, Goethe University Frankfurt \\
  \texttt{\{z.sun,s.gombert,f.zehner,l.camus,l.cong,h.drachsler\}@dipf.de}
}

\begin{document}
\maketitle
\begingroup
\renewcommand\thefootnote{}\footnotetext{All resources are available at \url{https://github.com/Szhifan/emnlp2026Ruspan}.}
\endgroup
\begin{abstract}
Automatic Short Answer Scoring (ASAS) requires models that can score student 
responses against question-specific criteria while remaining efficient and 
transferable across rubric sets.
We propose \textsc{Ruspan}, a rubric-conditioned 
ASAS framework that preserves a flexible natural-language rubric interface
while retaining single-pass inference.
\textsc{Ruspan} serializes the question context, 
student answer, and all candidate rubric levels into one sequence, 
then scores the levels listwise from rubric-span and whole-sequence 
representations produced in a single LM pass.
We further introduce \textsc{Ruspan}-RIM, 
where Rubric-Independent Mask prevents rubric spans from attending to one another, 
making rubric representations dependent only on the answer and question context and reducing 
dependence on rubric serialization patterns during zero-shot transfer.
On six ASAS benchmarks spanning English, German, and Portuguese, \textsc{Ruspan} is competitive with fixed-label and retrieval baselines in mono-benchmark scoring while remaining applicable to variable score ranges where fixed-label classifiers cannot be used, and RIM with position reindexing delivers consistent and substantial gains on \textsc{PT-ASAG}, the held-out benchmark with the strongest combined language and rubric-structure shift.
\end{abstract}
\section{Introduction}\label{sec:intro}

Automatic Short Answer Scoring (ASAS) is a central task in educational NLP,
offering the potential for timely and personalized feedback on student-constructed responses~\cite{asag_eras_trends, BaiStede2022SurveyAsas}.
Despite substantial progress, ASAS remains difficult to model uniformly across benchmarks.
Scoring schemes differ in the number, meaning, and structure of performance levels across questions and benchmarks.
Many early approaches therefore learn benchmark-specific label spaces: a static classifier is trained on top of a transformer encoder and its output dimensions correspond to the score levels of one dataset~\cite{kumar2019get, domain-pretrain-asas, riordan-etal-2017-investigating, ormerod2022shortanswerscoringensemblespretrained,asas-transformer}.
This approach is easy to implement and performs well within a benchmark, but it does not directly transfer when a new question or dataset requires a different scoring scheme.

A natural alternative is to use LLMs via prompting to directly score student answers,
either through commercial API calls or locally hosted models. 
However, commercial LLM APIs raise privacy and governance barriers for student writing~\cite{european_commission_gdpr_explained,uk_dfe2025_genai_education}, while locally hosted LLMs large enough for satisfactory zero- or few-shot performance are expensive to run.
Therefore, it is desirable to develop methods that are efficient at inference time yet retain the superior representation learning capabilities of LLMs.

In educational assessment, a rubric is a textual description that defines what constitutes each level of student performance.
Recent methods exploit this structure by framing ASAS as \textbf{rubric retrieval} or pairwise matching between rubric descriptions and example answers from each
level~\cite{bexte-etal-2022-similarity, sonkar2024automatedlonganswergrading}.
Such methods are computationally lighter than large generative models and flexible across benchmarks, as rubric descriptions encode the semantics of each score level rather than fixing a global label space.
However, these approaches typically expand each student answer into multiple answer-rubric pairs, one per candidate score level, increasing training and evaluation cost.

\citet{gombert2026rubric} present an early attempt at efficient rubric-conditioned scoring by 
encoding answers and rubrics together in a single sequence and aligning answer and rubric span
representations to predict scores via a bilinear network.
This avoids both fixed global label indices and pairwise expansion, thus offering a promising path to efficient and flexible ASAS\@.
We build on the insight that rubric spans can serve as label representations, but ask which contextual representation should accompany them, 
and how decoder-only LLMs can be adapted for discriminative scoring in this setting.
Our experiments show that pairing rubric span representations with the whole-sequence representation is substantially more effective than answer-rubric alignment.
Most importantly, although rubric retrieval and \citet{gombert2026rubric}'s proposed architecture both display 
potential for zero-shot transfer across datasets, neither has been systematically evaluated under zero-shot settings.

At the same time, recent work demonstrates that autoregressive LLMs can be repurposed as powerful discriminative text encoders, achieving strong NLU performance without generation~\cite{llm2vec,lookbothway,ruan2024largelanguagemodelsgoodclassifier,liu-etal-2024-rethink-llembed}.
This approach is appealing for ASAS because an LLM can represent and reason over long inputs containing the question, sample solution, student answer, and rubric criteria in one sequence.
Meanwhile, the rich world knowledge and the reasoning capabilities of LLMs are also beneficial for ASAS, which requires understanding of diverse subject matter and complex performance criteria.

Our contributions are as follows:
\begin{itemize}
    \item We propose \textsc{Ruspan}, a label-span encoding framework for rubric-conditioned ASAS\@. 
    It represents each score level through its rubric span representation and scores variable numbers of rubric levels in a single 
    forward pass using sequence-rubric alignment variants.
    \item We analyze a transfer limitation of standard causal attention: rubric representations become conditioned on preceding rubrics. We introduce Rubric-Independent Mask (RIM), a block-sparse attention masking mechanism that makes each rubric representation depend only on the answer and task context, approaching the rubric-set invariance of rubric retrieval while retaining single-pass efficiency.
    \item We evaluate \textsc{Ruspan} across six public ASAS benchmarks spanning multiple domains and languages against sequence classification, rubric retrieval, and LLM generation baselines. In mono-benchmark settings, \textsc{Ruspan} is competitive with the strongest single-function baselines while offering rubric flexibility over fixed-head classification and single-pass inference efficiency over retrieval and generation approaches. In cross-benchmark transfer, \textsc{Ruspan} with RIM delivers consistent and substantial gains under combined language and rubric-structure shift; on English-only cross-benchmark targets, retrieval baselines remain competitive.
\end{itemize}
\section{Related Work}\label{sec:related-work}
\subsection{Automatic Short Answer Scoring}\label{subsec:asas} 
Automatic Short Answer Scoring (ASAS) has evolved through several phases. 
Early work relied on handcrafted features such as lexical overlap with the reference answer and response length, 
combined with conventional machine learning models such as SVMs or linear regression~\cite{page1966imminence,madnani2013model, asag_eras_trends, dzikovska-etal-2013-semeval}. 
With the advent of pretrained transformers, ASAS research shifted toward neural approaches based on BERT-style encoders~\cite{asas-transformer, domain-pretrain-asas, ormerod2022shortanswerscoringensemblespretrained}, 
which typically formulate scoring as fixed-label classification with reference answers as the supporting context.
This design is effective when the scoring scheme is stable, yet scoring schemes can vary across questions and more strongly across benchmarks.
As a result, despite their ease of implementation, these methods often fail to generalize to new questions or datasets, sacrificing \textbf{flexibility} for within-benchmark \textbf{accuracy}.

To address this limitation, recent work has incorporated rubric information directly into the scoring process. In the context of ASAS, 
rubrics are textual descriptions of the performance criteria for each level. These descriptions provide 
semantic grounding, enabling models to evaluate student answers against explicit criteria rather than relying on implicit label representations.

One line of work frames scoring as rubric retrieval, where the model compares a student response with question-specific 
rubric descriptions and retrieves the most relevant one~\cite{sonkar2024automatedlonganswergrading,gombert2026rubric}.
A related approach uses reference or anchor answers from different score levels as implicit rubrics, 
scoring student answers through similarity-based comparison~\cite{bexte-etal-2022-similarity,li-etal-2023-answer-state}; because anchor answers encode 
the semantic traits expected at each level, they function analogously to textual rubric criteria.
Both formulations are more flexible than fixed-label classification as they rely 
on textual scoring criteria rather than a global label inventory.
However, they are usually implemented by expanding each student answer into multiple 
answer-rubric or answer-reference pairs,
which increases both training and inference cost, making it hard to scale to large datasets or models. This 
design trades off \textbf{efficiency} for \textbf{flexibility}, and it is not clear how well it performs in zero-shot transfer to other benchmarks. 

Recent work has also explored generative approaches using large language models 
(LLMs) for ASAS~\cite{chang2024finnish, chamieh2024llmasag, ferreiramello2025llmerasag, kortemeyer2024gpt4asag}. 
While predicting scores through text generation without assuming a fixed label space is \textbf{flexible}, it shifts scoring to a generative setup 
that is \textbf{inefficient} at inference time and sensitive to prompting and output formatting.

\subsection{LLMs as Encoders}
Recent research shows that decoder-only LLMs can serve as strong encoders for sentence embedding and sequence classification tasks
without autoregressive generation~\cite{llm2vec, llms-next-gen-dense-retrieval, lookbothway, lee2025nvembedimprovedtechniquestraining, qiao-etal-2025-decoderllmmaskedencoders, ruan2024largelanguagemodelsgoodclassifier, liu-etal-2024-rethink-llembed}.
This is appealing because it retains the representational capacity of LLMs while avoiding the cost and instability of autoregressive generation.
\citet{gekhman2025insideouthiddenfactualknowledge} further show that LLMs encode factual knowledge in their hidden states even when they fail to generate it reliably, 
suggesting that the encoder approach may be better suited to knowledge-intensive discrimination tasks such as ASAS\@.

However, prior work adapting LLMs as encoders has largely focused on single-sequence embedding and sentence-pair classification tasks.
Canonical evaluation benchmarks such as MTEB reflect this focus, as they contain mostly retrieval, semantic similarity, and NLI settings that involve at most two input sequences~\cite{muennighoff2023mtebmassivetextembedding, llm2vec, lee2025nvembedimprovedtechniquestraining, liu-etal-2024-rethink-llembed}.
The ability of LLMs to reason jointly over multiple heterogeneous inputs remains underexplored.
In ASAS, such multi-source reasoning is critical: a student answer must be evaluated against a question, a sample solution, and a set of rubric criteria.

\subsection{Label Semantics for NLU}
A broad line of work in NLU improves classification by incorporating label semantics directly into the input or prediction process. 
Examples include label verbalization~\cite{schick2021itsjustsizematters, hardalov2021fewshotcrosslingualstancedetection} 
and NLI-style formulations, where a label is expressed as a statement and 
evaluated against the text through a true/false judgment~\cite{xu-etal-2023-universal, elshabrawy-etal-2025-enabling}. 
These approaches replace purely index-based classification, 
which has been shown effective in zero- and few-shot settings. 
Rubric retrieval can be seen as a variant of this approach in ASAS, where each instance is expanded into multiple text-label pairs and
the model predicts whether a student answer satisfies the criteria described in each rubric.
\section{\textsc{Ruspan} Architecture and Benchmark Augmentation}\label{sec:method}
\subsection{Input Formulation}\label{subsec:input-formulation}
We formalize each benchmark as a dataset $\mathcal{D}=\mathopen{\{}(\mathbf{x}_i, y_i)\mathclose{\}}_{i=1}^{N}$, where $y_i$ is the ground-truth score level and the input instance is a tuple of textual fields:
\begin{equation}
\mathbf{x}_i = \textit{format}\big( q_i,\; s_i,\; a_i,\; \bar{r}_i,\; \iota_i \big).
\end{equation}
Here, $q_i$ is the question, $s_i$ is the sample solution, $a_i$ is the student answer, $\bar{r}_i$ denotes the concatenated rubric text,
and $\iota_i$ is the instruction sampled for this instance. We construct a pool of $M=10$ instruction templates and randomly sample one per instance during both training and testing, encouraging the model to be robust to instruction phrasing rather than overfitting to a fixed prompt format. For multilingual benchmarks, the instructions are written in the dataset language; the full pools are listed in \autoref{ap:multilingual}.
We obtain the final model input by serializing these fields into one sequence through the \textit{format} function.
Input examples can be found in \autoref{ap:input}.

\subsection{Rubric Span Extraction and Sequence Representation}\label{subsec:span-and-seq-repr}
\begin{figure*}[t]
    \centering
    \includegraphics[width=\textwidth]{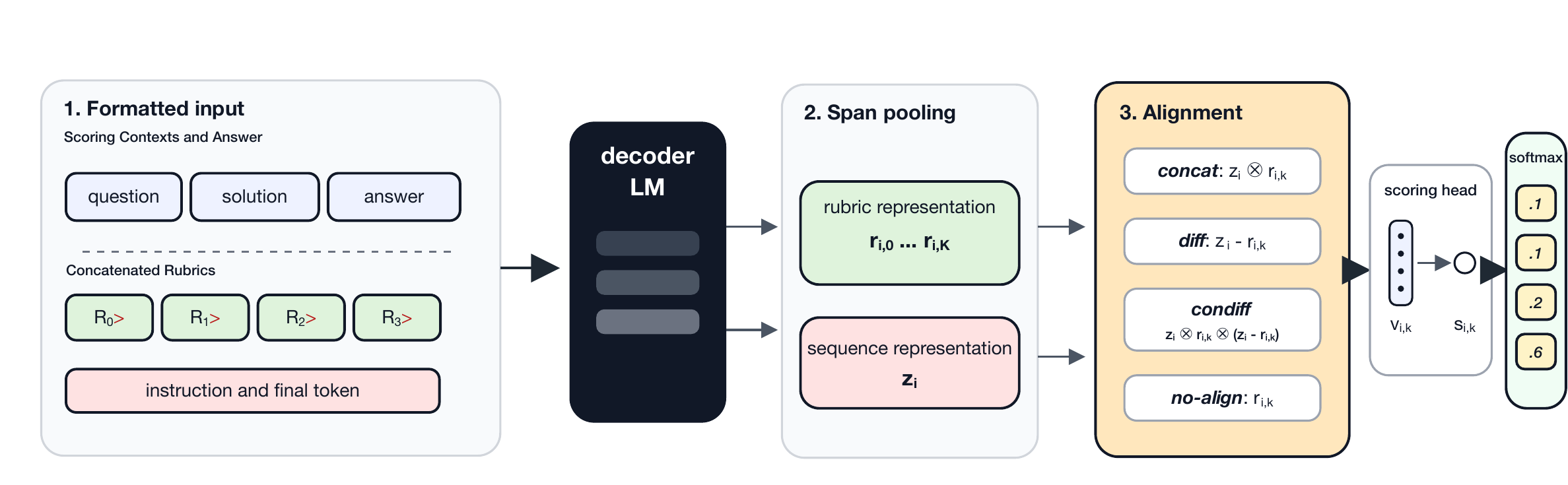}
    \caption{The \textsc{Ruspan} architecture. The formatted input combines the question context, student answer, rubric spans, and scoring instruction in one decoder-only LM pass. Span pooling extracts rubric representations from the final tokens of rubric spans and a sequence representation from the final token; an alignment function constructs each rubric-level representation before a shared scoring head projects it to a scalar score for softmax normalization.}\label{fig:method}
\end{figure*}

Given an input sequence $\mathbf{x}_i$, we keep track of rubric-span boundaries in the formatted text and, after language-model encoding, 
extract a compact representation for each rubric level by taking the hidden state of the last token of that rubric span. Let $\mathbf{H}_i\in\mathbb{R}^{T\times d}$ 
denote encoder outputs and let the $k$-th rubric span end at position $e_{i,k}$; we represent the rubric level 
as $\mathbf{r}_{i,k}=\mathbf{H}_i[e_{i,k}]$.
For global sequence representation, 
we use the final token embedding $\mathbf{z}_i=\mathbf{H}_i[T]$.
Unlike the \textit{answer-rubric} alignment of \citet{gombert2026rubric}, \textsc{Ruspan} pairs rubric spans with a
whole-sequence representation.
This final-token representation carries information from the full input in autoregressive models and is more effective than 
answer-rubric alignment alone, as demonstrated in \autoref{sec:results}.

\subsection{Label-to-Sequence Alignment Functions}\label{subsec:alignment-functions}
We now compute level scores by combining the rubric representations $\{\mathbf{r}_{i,k}\}$ extracted from the input $\mathbf{x}_i$ with the global sequence embedding $\mathbf{z}_i$ through alignment functions.
As shown in \autoref{fig:method}, we use four alignment functions. 
The first three combine $\mathbf{z}_i$ with each $\mathbf{r}_{i,k}$ 
into an aligned representation $\mathbf{v}_{i,k}$, 
while \textsc{no-align} uses the rubric representation directly:
\begin{itemize}
    \item \textsc{concat}: $\mathbf{v}_{i,k}=\mathbf{z}_i\otimes\mathbf{r}_{i,k}$
    \item \textsc{diff}: $\mathbf{v}_{i,k}=\mathbf{z}_i-\mathbf{r}_{i,k}$
    \item \textsc{condiff}: $\mathbf{v}_{i,k}=\mathbf{z}_i\otimes\mathbf{r}_{i,k}\otimes(\mathbf{z}_i-\mathbf{r}_{i,k})$
    \item \textsc{no-align}:  $\mathbf{v}_{i,k}=\mathbf{r}_{i,k}$
\end{itemize}
Here, $\otimes$ denotes vector concatenation.
Each resulting representation is projected to a scalar score $\hat{s}_{i,k}$ for level $k$ by a shared scoring head.

\subsection{Rubric-Independent Mask (RIM)}\label{subsec:rim}

Under standard causal attention, the hidden state of the final token of rubric $k$, denoted $\mathbf{r}_{i,k}$, attends to all tokens of rubrics $1, \ldots, k{-}1$
in addition to the context.
Consequently, each rubric representation is conditioned on the content
of the preceding rubrics. While learning contextual interactions between rubrics can boost performance in the mono-benchmark setting, this conditioning
degrades cross-benchmark transfer to unseen benchmarks with different rubric patterns, as shown in \autoref{sec:results}.

To remove this dependency while retaining the efficiency of a single forward pass, 
we introduce \textbf{Rubric-Independent Mask} (RIM), a block-sparse attention masking technique applied to standard causal attention.
Let the input sequence consist of a \emph{context block} (all tokens up to
and including the student answer) followed by $K$ \emph{rubric blocks}
spanning positions $s_k$ through $e_k$.

\begin{figure}[ht]
    \centering
    \includegraphics[width=1\linewidth]{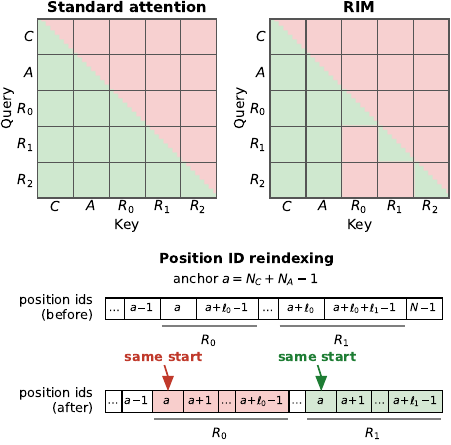}
    \caption{\textbf{Top}: Attention masks for standard autoregressive attention (left) vs.\ attention after applying RIM (right). Green = allowed; pink = blocked. Under RIM, each rubric block $R_k$ attends only to the context ($C$, $A$) and itself, never to other rubric blocks. \textbf{Bottom}: Position ID reindexing applied to each rubric span so that every rubric block starts its positional count from the same anchor offset ($N_C + N_A - 1$), preserving relative positions within each span.}\label{fig:rim}
\end{figure}

Under RIM, each rubric block attends only to the context and its own tokens,
never to other rubric blocks.
The resulting representation $\mathbf{r}_{i,k}$ is therefore a function of the
context and rubric $k$ alone.
We additionally evaluate a position-reindexed RIM variant, which shifts
the position IDs for each rubric block to start from the last token of the answer, so that the position embeddings
of each rubric token are also \textbf{independent} of other rubrics. The RIM and the position-reindexed variant are illustrated in \autoref{fig:rim}.
The instruction text $\iota_i$ is excluded from the RIM input because scoring no longer uses a final-token whole-sequence representation, 
as doing so would aggregate information across the whole serialized sequence and thereby reintroduce the rubric dependence that RIM is designed to remove.
Instead, each rubric representation $\mathbf{r}_{i,k}$ is passed directly to the scoring head using \textsc{no-align}. 

\subsection{Training Objective}\label{subsec:training-objective}
We optimize cross-entropy over gold labels and, since benchmarks differ in label cardinality and a batch may mix examples with different numbers of rubric levels, we pad each batch to the maximum and apply a validity mask $m_{i,k}\in\{0,1\}$ that sets padding-slot scores to $-\infty$ before normalization, so they do not affect the distribution.
This objective supports benchmark-specific label cardinalities while sharing the same encoder and scoring formulation.

\subsection{Benchmarks and LLM-Based Augmentation}\label{subsec:benchmarks-augmentation}
Because only two of the selected benchmarks contain native rubric annotations, we use LLMs to synthesize per-level rubric descriptions for the remaining datasets. 
This allows us to cast all benchmarks into a rubric-conditioned scoring formulation while preserving their original labels.

\subsubsection{Source Benchmarks}\label{subsubsec:source-benchmarks}
\begin{table*}[t]
\centering
\small
\resizebox{0.9\textwidth}{!}{%
\begin{tabular}{lllllll}
\toprule
Benchmark & Original Context & Augmented Context & Language & Levels & Test Splits & Sizes\\
\midrule
\textsc{Alice-LP}~\cite{sun2026alice}    & q, s, r & None & de & 3 & UA, UQ & 10,781 $|$ {--} $|$ 2,695 / 3,096 \\
\textsc{ASAP-SAS}    & q, r & s & en & 3--4 & UA & 17,043 $|$ {--} $|$ 5,224 \\
\textsc{iStudio}~\cite{li-etal-2023-answer-state}  & q, s & r & en & 3 & UA & 5,226 $|$ 653 $|$ 653 \\
\textsc{PT-ASAG}~\cite{galhardi2018portuguese}    & q, s & r & pt & 4 & UA & 3,290 $|$ {--} $|$ 366 \\
\textsc{SciEntsBank 3-way}~\cite{dzikovska-etal-2013-semeval} & q, s & r & en & 3 & UA, UQ, UD & 4,969 $|$ {--} $|$ 540 / 733 / 4,562 \\
\textsc{Beetle-3-way}~\cite{dzikovska-etal-2013-semeval}      & q, s & r & en & 3 & UA, UQ & 3,941 $|$ {--} $|$ 439 / 819 \\
\bottomrule
\end{tabular}}
\caption{Overview of ASAS benchmarks and their augmentation. \textit{q}: question; \textit{s}: sample solution; \textit{r}: rubrics. Test splits: \textit{UA} (unseen answers), \textit{UQ} (unseen questions), and \textit{UD} (unseen domains). Sizes are reported as train $|$ dev $|$ test; {--} indicates no split; multiple test sizes follow the order of the listed test splits.}\label{tab:benchmark}
\end{table*}

We collect six public ASAS benchmarks to evaluate \textsc{Ruspan}; \autoref{tab:benchmark} summarizes the datasets, 
language coverage, and available contextual information in each benchmark.
The original three-way setting of \textsc{SciEntsBank} and \textsc{Beetle}
uses the labels \textit{correct}, \textit{incorrect}, and \textit{contradictory}. In this work, we instead use the three-way variant \textit{correct}, \textit{partially correct}, and \textit{incorrect}, following the convention adopted by other ASAS benchmarks, 
and casting the remaining labels (\textit{contradictory}, \textit{irrelevant}, and \textit{non\_domain}) into \textit{incorrect}.

\subsubsection{Benchmark Augmentation}\label{subsubsec:rubric-augmentation}
For datasets without native rubric annotations, we use LLMs to synthesize per-level rubric descriptions. In addition, 
\textsc{ASAP-SAS} does not provide sample solutions, so we similarly use LLMs to generate sample solutions for this benchmark. The augmentation details are provided in \autoref{ap:augment}. 
\section{Experimental Setup} \label{section4}
Detailed hyperparameters and model choices are provided in Appendix~\ref{ap:hparams_model_choices}.

\subsection{Baselines}
We compare the proposed \textsc{Ruspan} method with the following baselines:
\paragraph{Sequence Classification}
We use a fixed-label classification head over the full input with rubric text appended. This baseline tests whether rubric text is sufficient when it is provided only as input context rather than represented as candidate score-level spans. \textsc{ASAP-SAS} is excluded because its number of score levels varies by question.

\paragraph{Rubric Retrieval}
We expand each instance into answer-rubric pairs, jointly encode each pair with an LM, 
and score whether the answer satisfies the rubric, 
following the implementation in \citet{sun2026alice}.

\paragraph{Answer-Rubric Alignment}
We compare our sequence-rubric alignment strategy with the answer-rubric alignment method proposed by \citet{gombert2026rubric}, which aligns the 
answer span representation with each rubric span representation through a bilinear scoring head.

\paragraph{Trained Generation via LLMs}
We fine-tune LLMs to generate the target score label.

\paragraph{Zero-shot Generation via LLMs}
We also evaluate frozen instruction-tuned LLMs that generate a score directly from the task context and rubric descriptions without task-specific fine-tuning. The prompt templates are provided in Appendix~\ref{ap:zero_shot_prompts}.

\subsection{Experimental Protocol}
We evaluate the effectiveness of \textsc{Ruspan} in two settings.

\paragraph{Mono-benchmark experiments}
We train and evaluate \textsc{Ruspan} and the baselines on each benchmark independently. For datasets with multiple
test splits, we report \textit{unseen question} and \textit{unseen domain} results when available, since they better test generalization beyond held-out answers. Zero-shot generation is excluded from this setting because it does not involve benchmark-specific fine-tuning.
For benchmarks without an official development split, we randomly sample 10\% of the training set as a held-out development set for checkpoint selection.

\paragraph{Joint training and cross-benchmark evaluation}
We jointly train a unified model on the English benchmarks, excluding \textsc{iStudio}. We report cross-benchmark results on \textsc{iStudio}, \textsc{PT-ASAG}, and \textsc{Alice-LP}.
\textsc{iStudio} tests same-language transfer ability,
while \textsc{PT-ASAG} and \textsc{Alice-LP} test transfer under stronger language or rubric shifts.
\section{Results and Discussion} \label{sec:results}

\subsection{Mono-Benchmark Results} \label{subsec:mono-results}

\autoref{tab:main_results}
reports macro-F1 on the ASAS benchmarks under the mono-benchmark fine-tuning setting, comparing \textsc{Ruspan} against sequence classification, answer-rubric alignment, rubric retrieval, and Trained Generation baselines.

\begin{table*}[t]
\centering
\small
\setlength{\tabcolsep}{4pt}
\resizebox{\textwidth}{!}{%
\begin{tabular}{ll ccccccc}
\toprule
\multirow{2}{*}{Model} & \multirow{2}{*}{Method}
    & \multicolumn{1}{c}{ASAP-SAS}
    & \multicolumn{1}{c}{Alice-LP}
    & \multicolumn{1}{c}{iStudio}
    & \multicolumn{1}{c}{PT-ASAG}
    & \multicolumn{2}{c}{SciEntsBank}
    & \multicolumn{1}{c}{Beetle} \\
\cmidrule(lr){3-3}\cmidrule(lr){4-4}\cmidrule(lr){5-5}\cmidrule(lr){6-6}\cmidrule(lr){7-8}\cmidrule(lr){9-9}
 & & UA & UQ & UA & UA & UQ & UD & UQ \\
\midrule
\multirow{5}{*}{\textit{Llama-3.2-1B-Instruct}}
    & Seq.\ Class.     & N/A           & 60.9 & 85.4          & 67.4          & 66.0\textsuperscript{$\dagger$} & 61.8\textsuperscript{$\dagger$} & 66.2 \\
    & Rub.\ Retrieval     & 74.3\textsuperscript{$\dagger$}          & 65.5          & 86.9 & 65.0          & 68.0 & 65.6\textsuperscript{$\dagger$} & \textbf{67.2} \\
    & Ans.\ Rub.\ Ali.    & 74.1\textsuperscript{$\dagger$} & 61.1\textsuperscript{$\dagger$} & 85.8          & 57.1\textsuperscript{$\dagger$} & 61.2\textsuperscript{$\dagger$} & 65.1\textsuperscript{$\dagger$} & 55.2\textsuperscript{$\dagger$} \\
    & Trained Gen.          & 56.1\textsuperscript{$\dagger$} & 46.5\textsuperscript{$\dagger$} & 83.5 & 54.9\textsuperscript{$\dagger$} & 44.5\textsuperscript{$\dagger$} & 44.9\textsuperscript{$\dagger$} & 59.4 \\
\cdashline{2-9}
    & \rule{0pt}{2.6ex}\textsc{Ruspan}    & 77.6-77.7-\textbf{77.8}-76.7 & 64.4-62.3-\textbf{66.5}-64.1 & \textbf{87.4}-85.9-86.6-85.9 & 66.0-63.8-\textbf{68.2}-63.1 & 65.6-65.2-64.2-\textbf{69.2} & \textbf{68.8}-66.5-68.0-66.7 & 66.4-64.9-61.9-65.7 \\
\midrule
\multirow{5}{*}{\textit{Llama-3.2-3B-Instruct}}
    & Seq.\ Class.     & N/A           & 64.6 & 87.5          & 67.7          & 70.8          & 67.8\textsuperscript{$\dagger$} & 66.0\textsuperscript{$\dagger$} \\
    & Rub.\ Retrieval     & 78.1\textsuperscript{$\dagger$}          & \textbf{69.5} & 87.8 & 66.4          & 71.3          & \textbf{72.3} & 69.5 \\
    & Ans.\ Rub.\ Ali.    & 76.4\textsuperscript{$\dagger$} & 62.8\textsuperscript{$\dagger$} & 87.4          & 70.2 & 69.7          & 69.4\textsuperscript{$\dagger$} & 61.1\textsuperscript{$\dagger$} \\
    & Trained Gen.          & 74.3\textsuperscript{$\dagger$} & 55.6\textsuperscript{$\dagger$} & 83.5\textsuperscript{$\dagger$} & 65.0          & 48.7\textsuperscript{$\dagger$} & 50.2\textsuperscript{$\dagger$} & 57.8\textsuperscript{$\dagger$} \\
\cdashline{2-9}
    & \rule{0pt}{2.6ex}\textsc{Ruspan}    & 80.2-79.7-\textbf{80.7}-78.7 & 64.5-65.3-66.5-65.2 & 88.6-87.9-86.8-\textbf{88.9} & \textbf{70.7}-69.7-69.3-68.7 & 71.6-69.7-\textbf{72.5}-70.5 & 70.3-71.7-71.8-71.5 & \textbf{69.7}-66.9-68.0-63.3 \\
\midrule
\multirow{4}{*}{\textit{Mistral-7B-v0.1}}
    & Seq.\ Class.     & N/A           & \textbf{68.9} & 88.3 & 69.0 & 66.0\textsuperscript{$\dagger$} & 67.6\textsuperscript{$\dagger$} & 67.0\textsuperscript{$\dagger$} \\
    & Ans.\ Rub.\ Ali.    & 75.5\textsuperscript{$\dagger$} & 64.2\textsuperscript{$\dagger$} & 86.7          & 62.1\textsuperscript{$\dagger$} & 72.3\textsuperscript{$\dagger$} & 69.4\textsuperscript{$\dagger$} & 61.5\textsuperscript{$\dagger$} \\
    & Trained Gen.          & 73.6\textsuperscript{$\dagger$} & 60.6\textsuperscript{$\dagger$} & 87.7          & 51.3\textsuperscript{$\dagger$} & 72.3\textsuperscript{$\dagger$} & 50.5\textsuperscript{$\dagger$} & 64.0\textsuperscript{$\dagger$} \\
\cdashline{2-9}
    & \rule{0pt}{2.6ex}\textsc{Ruspan}    & \textbf{80.8}-79.4-80.3-80.0 & 67.7-66.4-66.2-67.6 & 87.5-\textbf{88.4}-87.1-87.7 & 69.5-\textbf{70.3}-67.7-67.4 & 73.9-72.8-75.9-\textbf{76.2} & 70.9-71.6-71.4-\textbf{72.4} & \textbf{70.9}-69.1-68.0-66.9 \\
\midrule
\multirow{4}{*}{\textit{Llama-3.1-8B-Instruct}}
    & Seq.\ Class.     & N/A           & 67.2\textsuperscript{$\dagger$} & 88.8          & 65.6          & 68.6\textsuperscript{$\dagger$} & 68.3\textsuperscript{$\dagger$} & 66.3\textsuperscript{$\dagger$} \\
    & Ans.\ Rub.\ Ali.    & 78.8 & 62.6\textsuperscript{$\dagger$} & 87.8          & 64.5\textsuperscript{$\dagger$} & 72.2          & 72.6\textsuperscript{$\dagger$} & 64.4\textsuperscript{$\dagger$} \\
    & Trained Gen.          & 74.6\textsuperscript{$\dagger$} & 62.5\textsuperscript{$\dagger$} & 86.3\textsuperscript{$\dagger$} & 58.2\textsuperscript{$\dagger$} & 70.6\textsuperscript{$\dagger$} & 71.5\textsuperscript{$\dagger$} & 64.8\textsuperscript{$\dagger$} \\
\cdashline{2-9}
    & \rule{0pt}{2.6ex}\textsc{Ruspan}    & 78.7-79.3-78.7-\textbf{79.4} & 67.3-68.4-\textbf{69.6}-67.5 & 89.0-89.0-\textbf{89.8}-88.1 & 64.3-67.4-68.3-\textbf{69.9} & 70.2-\textbf{75.1}-74.7-70.1 & 70.4-70.4-\textbf{74.1}-73.2 & 66.5-\textbf{70.5}-68.0-68.6 \\
\bottomrule
\end{tabular}}
\caption{
    Main mono-benchmark results (macro-F1) on ASAS benchmarks.
    For \textsc{Ruspan}, hyphen-separated scores are ordered as \textsc{concat}-\textsc{diff}-\textsc{condiff}-\textsc{no-align}.
    \textbf{Bold} indicates the best score within each backbone and benchmark column.
    A baseline score marked with \textsuperscript{$\dagger$} indicates a significant one-sided paired approximate-randomization improvement of the best \textsc{Ruspan} variant over that baseline ($p<.05$).
    Rubric Retrieval is omitted for 7B and 8B models due to high computational cost.
}
\label{tab:main_results}
\end{table*}

In \autoref{tab:main_results}, at least one \textsc{Ruspan} alignment variant is the strongest method on \textsc{ASAP-SAS} across all four backbones and achieves the best or near-best result in several cells for the 1B and 8B backbones.
At the same time, \colorbox{gray!15}{Sequence Classification} and \colorbox{gray!15}{Rubric Retrieval} are also competitive, and both are stronger than \textsc{Ruspan} in several individual benchmark and backbone combinations.
Regarding model scaling, the 1B backbone is the weakest overall and lags notably behind larger models, whereas gains from scaling diminish going from 3B to 8B.
The flexibility advantage is especially clear on benchmarks with variable score spaces such as \textsc{ASAP-SAS}: 
because \textsc{Ruspan} scores the candidate rubric spans directly, it remains applicable when fixed-label \colorbox{gray!15}{Sequence Classification} is not.

Compared with the \colorbox{gray!15}{Answer-Rubric Alignment} method in \citet{gombert2026rubric},
\textsc{Ruspan}'s whole-sequence conditioning is more effective in most settings in \autoref{tab:main_results}.
\textsc{Ruspan} is also usually stronger than \colorbox{gray!15}{Trained Generation}
in mono-benchmark ASAS; for example,
with the 1B backbone on \textsc{Alice-LP}, \textsc{Ruspan}
reaches 66.5 macro-F1 compared with 46.5 for \colorbox{gray!15}{Trained Generation}.

The alignment-function breakdown in \autoref{tab:main_results} shows that while no single alignment function is universally optimal across benchmarks and backbone sizes, \textsc{condiff} performs generally well.
Nevertheless, \textsc{Ruspan} is generally competitive across alignment choices: when one alignment underperforms in a given setting, another often closes the gap or exceeds the strongest baseline.
We therefore interpret the mono-benchmark comparison as an evaluation of the \textsc{Ruspan} alignment family rather than evidence for a single selected alignment function.

\subsection{Joint Training and Cross-Benchmark Results} \label{subsec:multi-results}

\autoref{tab:multi_results}
reports macro-F1 on the cross-benchmark evaluation sets under joint training on the English ASAS benchmarks.
For \textsc{Ruspan}, we fix the alignment function to \textsc{condiff}.

\begin{table}[t]
\centering
\small
\setlength{\tabcolsep}{3pt}
\resizebox{\columnwidth}{!}{%
\begin{tabular}{ll ccc}
\toprule
\multirow{2}{*}{Model} & \multirow{2}{*}{Method}
    & \multicolumn{1}{c}{\textit{iStudio}}
    & \multicolumn{1}{c}{Alice-LP}
    & \multicolumn{1}{c}{PT-ASAG} \\
\cmidrule(lr){3-3}\cmidrule(lr){4-4}\cmidrule(lr){5-5}
 & & UA & UQ & UA \\
\midrule
\multirow{6}{*}{\textit{Llama-3.2-1B-Instruct}}
    & Rub.\ Retrieval                    & 54.4{\tiny$\pm$2.4}           & 45.5{\tiny$\pm$0.7}           & 38.2{\tiny$\pm$2.7}           \\
    & Trained Gen.                       & 57.6{\tiny$\pm$3.1}           & \textbf{48.8}{\tiny$\pm$2.6}  & 30.6{\tiny$\pm$1.7}           \\
    & Zero-shot Gen.                     & 27.7                          & 44.7                          & 32.8                          \\
\cdashline{2-5}
    & \textsc{Ruspan}                    & 55.8{\tiny$\pm$5.5}           & 48.4{\tiny$\pm$2.1}           & 28.1{\tiny$\pm$0.6}           \\
    & \textsc{Ruspan}-RIM                & \textbf{59.9}{\tiny$\pm$2.6}  & 42.9{\tiny$\pm$7.8}           & 29.9{\tiny$\pm$3.3}           \\
    & \textsc{Ruspan}-RIM-reindexing     & 57.6{\tiny$\pm$8.1}           & 41.9{\tiny$\pm$1.2}           & \textbf{40.6}{\tiny$\pm$0.6}  \\
\midrule
\multirow{6}{*}{\textit{Llama-3.2-3B-Instruct}}
    & Rub.\ Retrieval                    & 68.1{\tiny$\pm$5.3}           & 53.4{\tiny$\pm$5.0}           & \textbf{47.3}{\tiny$\pm$2.2}  \\
    & Trained Gen.                       & \textbf{71.1}{\tiny$\pm$2.6}  & 45.9{\tiny$\pm$6.9}           & 33.5{\tiny$\pm$0.6}           \\
    & Zero-shot Gen.                     & 21.9                          & 42.3                          & 24.8                          \\
\cdashline{2-5}
    & \textsc{Ruspan}                    & 65.3{\tiny$\pm$4.6}           & 53.8{\tiny$\pm$2.0}           & 34.8{\tiny$\pm$3.3}           \\
    & \textsc{Ruspan}-RIM                & 70.9{\tiny$\pm$2.4}           & \textbf{56.5}{\tiny$\pm$4.1}  & 41.2{\tiny$\pm$0.8}           \\
    & \textsc{Ruspan}-RIM-reindexing     & 67.9{\tiny$\pm$4.5}           & 53.2{\tiny$\pm$1.7}           & 46.3{\tiny$\pm$2.2}           \\
\midrule
\multirow{5}{*}{\textit{Mistral-7B-v0.1}}
    & Trained Gen.                       & \textbf{70.0}{\tiny$\pm$1.8}  & 55.3{\tiny$\pm$5.5}           & 35.7{\tiny$\pm$2.4}           \\
    & Zero-shot Gen.                     & 61.2                          & 49.2                          & \textbf{45.6}                 \\
\cdashline{2-5}
    & \textsc{Ruspan}                    & 67.8{\tiny$\pm$8.0}           & 50.7{\tiny$\pm$3.4}           & 32.0{\tiny$\pm$0.6}           \\
    & \textsc{Ruspan}-RIM                & 69.2{\tiny$\pm$4.0}           & \textbf{56.1}{\tiny$\pm$2.2}  & 40.6{\tiny$\pm$2.0}           \\
    & \textsc{Ruspan}-RIM-reindexing     & 67.4{\tiny$\pm$6.4}           & 54.9{\tiny$\pm$0.4}           & 44.0{\tiny$\pm$3.3}           \\
\midrule
\multirow{5}{*}{\textit{Llama-3.1-8B-Instruct}}
    & Trained Gen.                       & 71.2{\tiny$\pm$3.7}           & 39.6{\tiny$\pm$4.1}           & 35.1{\tiny$\pm$2.3}           \\
    & Zero-shot Gen.                     & 49.1                          & 55.3                          & 36.4                          \\
\cdashline{2-5}
    & \textsc{Ruspan}                    & 66.5{\tiny$\pm$3.6}           & 53.9{\tiny$\pm$1.3}           & 41.3{\tiny$\pm$4.2}           \\
    & \textsc{Ruspan}-RIM                & 70.7{\tiny$\pm$4.3}           & 54.8{\tiny$\pm$0.8}           & 46.9{\tiny$\pm$1.8}           \\
    & \textsc{Ruspan}-RIM-reindexing     & \textbf{71.3}{\tiny$\pm$3.2}  & \textbf{56.7}{\tiny$\pm$1.4}  & \textbf{48.2}{\tiny$\pm$1.1}  \\
\bottomrule
\end{tabular}}
\caption{%
    Joint-training cross-benchmark macro-F1 (mean {\tiny$\pm$ std} over 3 random seeds).
    \textbf{Bold}: best method within each model group.
    Zero-shot Gen.\ results are from a single run.
}
\label{tab:multi_results}
\end{table}

\paragraph{Standard \textsc{Ruspan} transfers unevenly across benchmarks.}
The joint-training cross-benchmark setting is substantially harder than mono-benchmark fine-tuning for standard \textsc{Ruspan}.
Transfer is especially difficult for \textsc{PT-ASAG}, followed by \textsc{Alice-LP} and \textsc{iStudio}, as 
the training benchmarks are English and mostly use three-level score scales, except for \textsc{ASAP-SAS}, which includes three- and four-level questions.

On \textsc{Alice-LP}, standard \textsc{Ruspan} remains broadly competitive 
with the cross-benchmark baselines, matching or exceeding \colorbox{gray!15}{Trained Generation} for the 3B and 8B backbones.
On \textsc{iStudio}, the strongest \textsc{Ruspan} variants are competitive, while standard \textsc{Ruspan} is more variable across model sizes.
However, for \textsc{PT-ASAG}, standard \textsc{Ruspan} falls well behind the baselines. 
\paragraph{Inter-rubric dependence contributes to cross-benchmark sensitivity.}
These results are consistent with the diagnosis in \autoref{subsec:rim}, and two independent pieces of evidence point in the same direction.
Standard \textsc{Ruspan} degrades most on \textsc{PT-ASAG}, where rubric structure and language both shift strongly from the training pool.
\colorbox{gray!15}{Rubric Retrieval}, which scores each answer-rubric pair independently
and is therefore structurally immune to inter-rubric attention dependence,
maintains robust performance on \textsc{PT-ASAG}.
\textsc{Ruspan}-RIM variants enforce the same per-rubric independence through attention masking, and the position-reindexed variant is consistently stronger than standard \textsc{Ruspan} on \textsc{PT-ASAG}.
The convergence of two architecturally distinct
approaches that share per-rubric independence
as a design property is consistent
with rubric representation decoupling as an important factor
associated with cross-benchmark robustness.

\paragraph{RIM with reindexing delivers consistent gains under rubric structure shifts.}
\textsc{Ruspan}-RIM with position reindexing improves over standard \textsc{Ruspan} on \textsc{PT-ASAG} across all four backbone sizes, with gains ranging from 6.9 to 12.5 macro-F1 points.
This uniformity is notable: \textsc{PT-ASAG} combines two strong sources of distribution shift, Portuguese text and a four-level score scale, yet the improvement holds regardless of backbone size.

\textsc{Ruspan}-RIM is competitive on all three transfer targets rather than useful only under strong shift.
Taking the better of the two RIM variants in each cell, RIM matches or exceeds the strongest baseline in 7 of the 12 benchmark-backbone combinations, and in four of the remaining five it trails by at most 1.6 macro-F1.
Relative to standard \textsc{Ruspan}, RIM improves in 11 of 12 combinations.
The \emph{size} of the gain also tracks the severity of the shift: largest on \textsc{PT-ASAG} (up to $+12.5$), moderate on \textsc{iStudio} and \textsc{Alice-LP}.
With the 8B backbone, RIM with position reindexing is the strongest method on all three targets.

\paragraph{Accuracy, flexibility, and efficiency tradeoffs.}
\autoref{tab:method_comparison} summarizes how the evaluated methods differ in label-space flexibility, input processing, prediction mechanism, and task-specific training.

\begin{table*}[t]
\centering
\small
\setlength{\tabcolsep}{5pt}
\renewcommand{\arraystretch}{1.08}
\resizebox{0.7\textwidth}{!}{%
\begin{tabular}{@{}l c c p{0.29\textwidth} c@{}}
\toprule
\textbf{Method} & \textbf{Variable $K$} & \textbf{Forward-pass per instance} & \textbf{Prediction mechanism} & \textbf{Task-trained} \\
\midrule
Seq. Classification
    & No  & 1 & Fixed $K$-way classification head & Yes \\
Rubric Retrieval
    & Yes & K & Expand each instance into answer-rubric pairs and take the rubric with the highest score & Yes \\
Trained Generation
    & Yes & 1 & Autoregressive score generation & Yes \\
Zero-shot Generation
    & Yes & 1 & Autoregressive score generation & No \\
\midrule
\textsc{Ruspan}
    & Yes & 1 & $K$ parallel rubric-span scores & Yes \\
\bottomrule
\end{tabular}
}
\caption{Structural comparison of ASAS methods. $K$ denotes the number of available levels. Variable $K$ indicates whether one trained model can handle different numbers of candidate levels.}
\label{tab:method_comparison}
\end{table*}

\begin{figure}[t]
    \centering
    \includegraphics[width=\columnwidth]{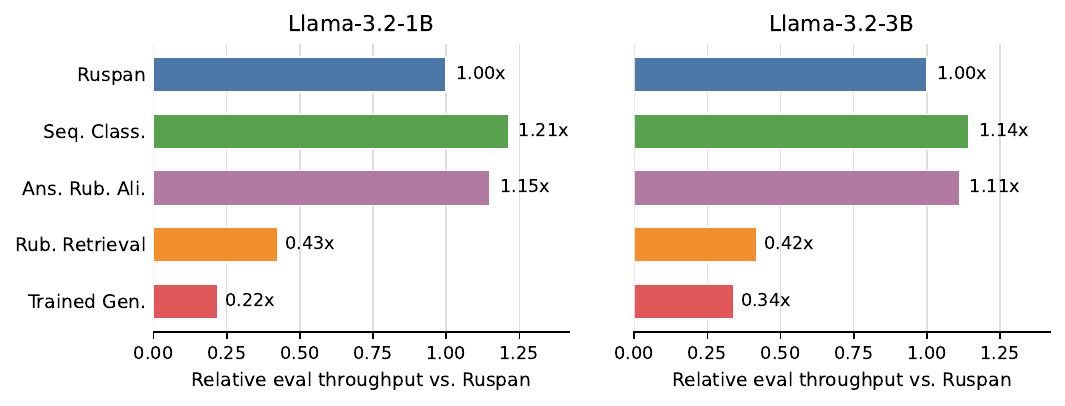}
    \caption{Relative evaluation throughput on \textsc{iStudio}, normalized by \textsc{Ruspan} within each model size. Values above 1.0 indicate faster inference than \textsc{Ruspan}; values below 1.0 indicate slower inference.}
    \label{fig:eval_speed_vs_ruspan}
\end{figure}

\autoref{fig:eval_speed_vs_ruspan} shows throughput on \textsc{iStudio}, normalized by \textsc{Ruspan}.
Fixed-head baselines are faster, while \colorbox{gray!15}{Rubric Retrieval} and \colorbox{gray!15}{Trained Generation} are substantially slower.\footnote{\colorbox{gray!15}{Zero-shot Gen.}\ is excluded because it uses vLLM, whereas all other methods use the Hugging Face Trainer evaluation loop.}
\textsc{Ruspan} thus sits between these extremes: modestly slower than fixed-head classifiers but much faster than retrieval and generation methods, while retaining competitive accuracy in mono-benchmark settings,
and achieves substantial cross-benchmark generalization performance with RIM and reindexing.

\subsection{Per-Level Analysis} \label{subsec:per_level_analysis}

\begin{figure*}[t]
\centering
\includegraphics[width=0.68\textwidth]{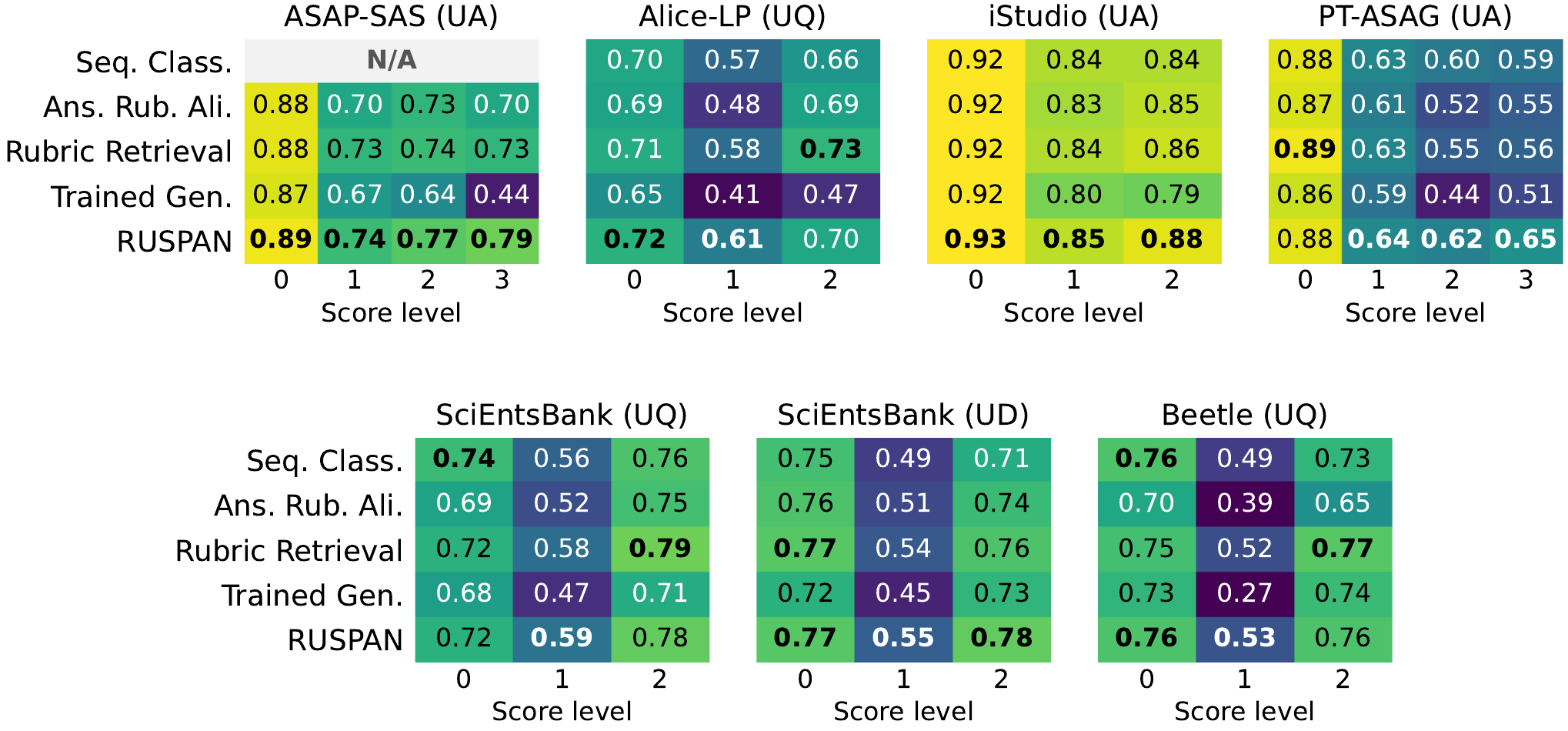}\\[1pt]
\rule{0.68\textwidth}{0.35pt}\\[-1pt]
\includegraphics[width=0.68\textwidth]{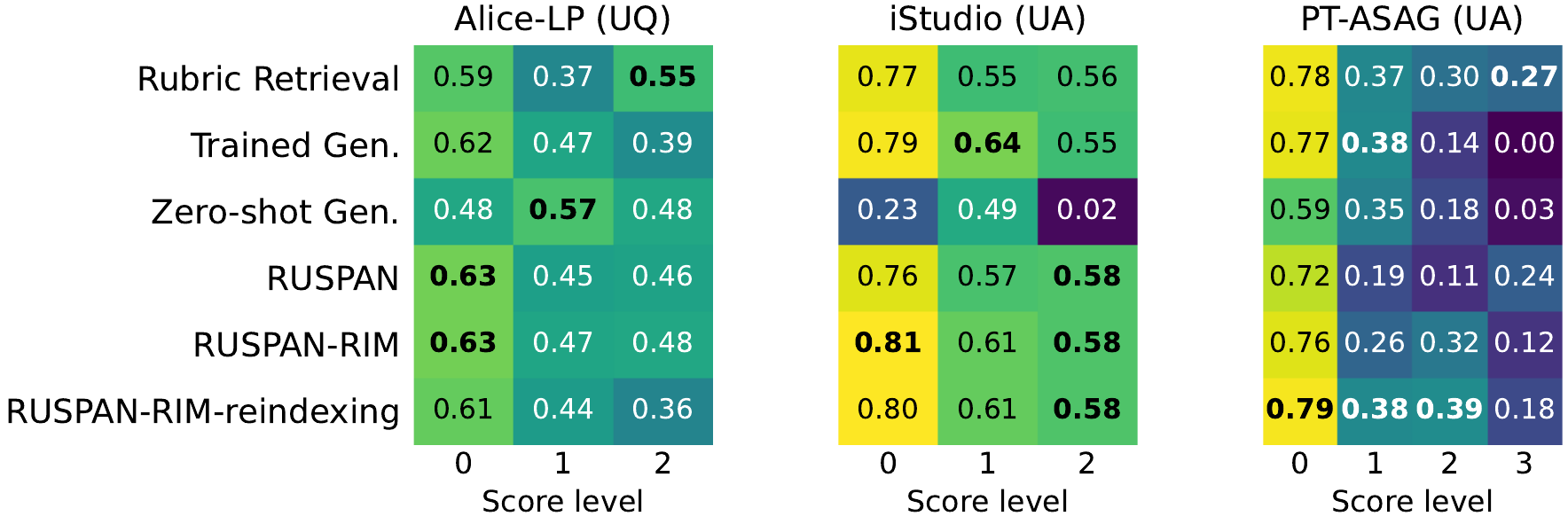}
\caption{Per-level F1 by method. Top: mono-benchmark training. Bottom: cross-benchmark, averaged over random seeds. \textbf{Bold} marks the best result at each score level.}
\label{fig:per_level_results}
\end{figure*}

\autoref{fig:per_level_results} compares per-level F1 under mono-benchmark and cross-benchmark training.
We include only the 1B and 3B backbones because the 7B and 8B backbones are excluded from \colorbox{gray!15}{Rubric Retrieval}.
Training-set label distributions are shown in \autoref{fig:label_distribution}.

\paragraph{Label imbalance partially explains level gaps.}
Majority or near-majority levels consistently receive higher F1, while rare levels remain harder even when overall macro-F1 is strong.
For example, level~3 in \textsc{PT-ASAG} has the smallest support, roughly 13\% of training examples, and all methods score it well below the majority level~0.
This indicates that part of the per-level gap reflects data imbalance rather than model architecture.

\paragraph{Intermediate levels are the hardest under balanced distributions.}
When label distributions are roughly balanced, as in \textsc{Beetle}, \textsc{SciEntsBank}, and \textsc{Alice-LP}, intermediate levels consistently yield the lowest F1 across all methods.
This is consistent with semantic overlap between adjacent categories being hardest to resolve, independently of frequency.

\paragraph{Trained Generation exhibits recency bias.}
The Trained Generation baseline degrades more sharply on higher score levels than other methods~\citep{yu2025mitigatepositionbiaslarge}.
Because rubrics are presented in ordinal order, the highest score level appears last in the input sequence, making generation-based prediction sensitive to the final rubric position.
The effect is especially pronounced on \textsc{Beetle}, with level~1 F1 of 0.27 compared with 0.53 for \textsc{Ruspan}, and on \textsc{Alice-LP}.

\paragraph{\textsc{Ruspan} performs best on the hardest levels.}
For \textsc{SciEntsBank}, \textsc{Beetle}, and \textsc{Alice-LP}, where intermediate levels are hardest, \textsc{Ruspan} achieves the highest or near-highest level~1 F1 among all methods.
For \textsc{PT-ASAG}, where the hardest levels are the higher-score levels~2 and~3, \textsc{Ruspan} again leads on those levels.
This suggests that aligning predictions to individual rubric spans provides a more reliable signal for ambiguous score boundaries than prompt-level classification or retrieval.

Under cross-benchmark training, the gain from \textsc{Ruspan}-RIM with position reindexing on \textsc{PT-ASAG} is prominent on the harder levels.
Compared with standard \textsc{Ruspan}, the reindexed RIM variant improves F1 on all four levels, with the largest relative gains on the harder higher-score levels.
This pattern supports the interpretation that reducing rubric-order and rubric-cardinality dependence is most useful under the combined language and four-level rubric shift in \textsc{PT-ASAG}.

\paragraph{Further analyses.}
\autoref{ap:ablation} tests rubric detail, human-written versus LLM-generated rubrics, and the contribution of the LLM-generated sample solution on \textsc{ASAP-SAS}.
\autoref{ap:non_asas} evaluates generalization beyond ASAS on seven NLU benchmarks spanning commonsense, stance, figurative language, sentiment, topic, and scientific edit intent.

\section{Conclusion} \label{sec:conclusion}

We introduced \textsc{Ruspan}, a rubric-conditioned framework for short answer scoring that represents score levels as rubric spans jointly encoded with the student answer.
By scoring all candidate levels in a single forward pass, \textsc{Ruspan} provides a flexible and efficient alternative within the tradeoff among flexibility, accuracy, and efficiency that limits prior ASAS approaches, including variable score spaces where fixed-head classifiers cannot be used.
Across six ASAS benchmarks, \textsc{Ruspan} delivers competitive mono-benchmark performance while avoiding pairwise answer-rubric expansion and autoregressive score generation.
For cross-benchmark transfer, RIM with position reindexing mitigates inter-rubric attention dependence, yielding consistent and substantial gains on \textsc{PT-ASAG} under combined language and rubric-structure shift.
Future work should explore richer feedback generation, multi-trait scoring, and robustness to noisy or underspecified rubrics.

\section{Limitations}

A central limitation of this work is the quality of the rubric annotations used for benchmark augmentation. 
We emphasize that rubric generation is not the object of study in this paper. 
Rather, generated rubrics serve as a practical augmentation step that allows benchmarks without native rubric descriptions 
to be evaluated under a common rubric-conditioned formulation. 
Our claims therefore concern the model's ability to use natural-language label descriptions as score-level representations, 
not the pedagogical optimality of the generated rubrics. 
In real educational settings, rubrics are typically written by trained 
educators who have access to curriculum goals, instructional materials, grading policies, 
and broader pedagogical context. 

However, this information is not usually available in public ASAS datasets. 
This limited context can produce rubric descriptions that are under-specified, 
overly generic, or misaligned with the intended underlying assessment criteria. 

Our study is also limited by the scope of public ASAS benchmarks. These datasets provide 
controlled short-answer scoring tasks, but they do not fully capture the complexity of classroom assessment,
where responses may require multi-trait scoring, richer feedback, or reasoning over longer and more diverse student work. 
Finally, cross-benchmark transfer remains uneven, especially under language and rubric-structure shifts.
While RIM improves robustness in several settings, preventing interactions among rubric spans may also remove useful information when score levels are meaningfully related.

\section*{Acknowledgments}

This work was supported by the Volkswagen Foundation through the project \emph{From Machine Learning to Machine Teaching: Making Machines AND Humans Smarter} (ML2MT).
The GPU resources used in this work were provided by Hessian.ai.

\bibliography{custom}

\appendix
\raggedbottom

\section{Benchmark Augmentation} \label{ap:augment}

Several benchmarks in our collection lack rubric descriptions or reference answers. We augment these benchmarks using \texttt{gpt-4o-mini} via two pipelines described below, with the prompts shown in Figures~\ref{fig:ref_ans_prompt} and~\ref{fig:rubric_prompt}.

\paragraph{Reference Answer Generation.}
\textsc{ASAP-SAS} does not provide reference answers. For each question, we prompt the LLM with the question text to generate a reference answer that serves as the sample solution during scoring (Figure~\ref{fig:ref_ans_prompt}).

\paragraph{Rubric Generation.}
For benchmarks without rubric descriptions, we prompt the LLM to generate a 2--3 sentence rubric for each score level (Figure~\ref{fig:rubric_prompt}). The prompt supplies the question text, all available reference answers, up to 10 student answer examples per score level from the training split, and the set of target score labels. The model is instructed to describe responses at each level directly (e.g., ``Responses demonstrate\ldots'') without referencing label names, and to write in the language of the dataset.
The LLM-generated rubric descriptions and sample solutions are derived from publicly available research datasets and are intended for research use only.

\begin{figure}[h]
\centering
\small
\begin{lstlisting}[
    basicstyle=\ttfamily\small,
    breaklines=true,
    frame=single,
    columns=fullflexible,
    keepspaces=true
]
Generate a high-quality reference answer for the
following question. The answer should be an
independent, complete, and concise response that
can serve as a scoring reference.

Question:
[QUESTION]

Requirements:
- The answer should correctly address the question;
- Each answer should be only ONE sentence long;
- Ensure the language of the answer matches the
  language of the question.
\end{lstlisting}
\caption{Prompt template for reference answer generation.}
\label{fig:ref_ans_prompt}
\end{figure}

\begin{figure}[h]
\centering
\small
\begin{lstlisting}[
    basicstyle=\ttfamily\small,
    breaklines=true,
    frame=single,
    columns=fullflexible,
    keepspaces=true
]
### System
You are an expert educator specializing in
assessment and rubric development. Your task is
to generate clear, criterion-based rubrics that
distinguish between different score categories.
The score labels for this question are: [LABELS]
CRITICAL: Do not mention the label, category,
level, or score name inside the rubric text.
Describe responses directly using phrasing like
"Responses ..." in the dataset language.

### User
Question:
[QUESTION]

Reference Answer(s):
[REFERENCE ANSWERS]

Student Response Examples by Score Level:
[LEVEL] Examples:
  1. [ANSWER]
  ...

For EACH score level, write a 2-3 sentence rubric
that explains:
1. What knowledge the responses demonstrate;
2. What characteristics distinguish this level;
3. Clear criteria for consistent categorization.

Return as JSON:
{
  "rubrics": {"[LEVEL]": "...", ...},
  "reasoning": "..."
}
\end{lstlisting}
\caption{Prompt template for rubric generation.}
\label{fig:rubric_prompt}
\end{figure}

\begin{figure}[h]
\centering
\small
\noindent\fbox{\parbox{0.95\linewidth}{
\textbf{Question:} Explain why you got a voltage reading of 1.5 for terminal 1 and the positive terminal.

\smallskip
\textbf{Sample Solution:} Terminal 1 and the positive terminal are separated by a gap, so the voltmeter measures the full battery voltage of 1.5\,V across the open circuit.

\smallskip
\textbf{Student Answers by Score Level:}
\begin{itemize}
  \item[\textbf{0}] \textit{``because terminal one and the positive terminal are connected''}
  \item[\textbf{1}] \textit{``because there was a gap in the connection''}
  \item[\textbf{2}] \textit{``because there is a gap between terminal one and the positive battery terminal''}
\end{itemize}

\smallskip
\textbf{LLM-Generated Rubrics:}
\begin{itemize}
  \item[\textbf{0}] Responses demonstrate a basic recognition of the connection between terminal~1 and the positive terminal but lack clarity and detail. They assert that the terminals are connected without explaining the implications of this connection or providing context about the voltage reading.
  \item[\textbf{1}] Responses indicate some understanding of separation or gaps between terminals but fail to accurately connect this understanding to the voltage reading. There is confusion about the relationship between the terminals, often mentioning separation without a clear explanation of its relevance.
  \item[\textbf{2}] Responses clearly articulate the concept of a gap or separation between terminal~1 and the positive battery terminal, demonstrating an understanding of how this affects the voltage reading. They accurately describe the relationship between the terminals, indicating an awareness of the electrical principles involved.
\end{itemize}
}}
\caption{Example of LLM-generated rubrics for a question from the \textsc{Beetle} dataset. Score levels range from 0 (incorrect) to 2 (correct).}
\label{fig:rubric_example}
\end{figure}

\FloatBarrier

\section{Input Examples} \label{ap:input}

We follow \citet{ruan2024largelanguagemodelsgoodclassifier} in using XML-style tags to mark field boundaries.

\begin{figure}[h]
\centering
\small
\begin{lstlisting}[
    basicstyle=\ttfamily\small,
    breaklines=true,
    frame=single,
    columns=fullflexible,
    keepspaces=true,
    escapeinside={(*@}{@*)}
]
<question>
Starting with mRNA leaving the nucleus,
list and describe four major steps
involved in protein synthesis.
</question>
<sample solution>
The first step in protein synthesis
involves the mRNA exiting the nucleus
and entering the cytoplasm, where it
serves as a template for protein
assembly.
</sample solution>
<answer>
mRNA leaves the nucleus mRNA carries
the instruction to the mitocondria
mitocondria then makes the protiens
at last the mRNA returns to the nucleus
with the protiens
</answer>
(*@\colorbox{cyan!20}{\texttt{<rubric> Other </rubric}}\colorbox{red!20}{\texttt{>}}@*)
(*@\colorbox{cyan!20}{\texttt{<rubric> One or two key elements </rubric}}\colorbox{red!20}{\texttt{>}}@*)
(*@\colorbox{cyan!20}{\texttt{<rubric> Three key elements </rubric}}\colorbox{red!20}{\texttt{>}}@*)
(*@\colorbox{cyan!20}{\texttt{<rubric> Four key elements </rubric}}\colorbox{red!20}{\texttt{>}}@*)
Evaluate the answer based on the
provided rubric and other context
information.
\end{lstlisting}
\caption{An example model input from \textsc{ASAP-SAS}, showing the question, sample solution, student answer, candidate rubric labels (highlighted in blue), and scoring instruction. The \colorbox{red!20}{\texttt{>}} token at the end of each rubric span is used as the pooling token to extract the rubric representation.}
\label{fig:input_example}
\end{figure}

\FloatBarrier

\section{Multilingual Input Format and Instruction Pool} \label{ap:multilingual}

For multilingual benchmarks we translate both the XML field tags and the scoring instructions into the dataset language. \autoref{tab:field_translations} shows the field tag mappings for German (\texttt{alice\_lp}) and Portuguese (\texttt{pt\_asag}). The \texttt{<rubric>} tag is kept in English across all benchmarks as it is an internal structural marker.
We use the multilingual format in mono-benchmark training and evaluation. 
For multi-benchmark training, we use the English format for all benchmarks, 
as the models are trained on the English format.

\begin{table}[h]
\centering
\small
\resizebox{0.9\linewidth}{!}{%
\begin{tabular}{lccc}
\toprule
Field & English & German & Portuguese \\
\midrule
question      & \texttt{<question>}       & \texttt{<Frage>}          & \texttt{<pergunta>} \\
sample solution & \texttt{<sample solution>}& \texttt{<Musterlösung>}   & \texttt{<solução\_exemplar>} \\
answer        & \texttt{<answer>}         & \texttt{<Antwort>}        & \texttt{<resposta>} \\
rubric        & \texttt{<rubric>}         & \texttt{<rubric>}         & \texttt{<rubric>} \\
\bottomrule
\end{tabular}
}
\caption{XML field tag translations by language.}
\label{tab:field_translations}
\end{table}

We compose a pool of $M=10$ scoring instruction templates per language; one is randomly sampled per instance at training and test time. The full pools are listed in \autoref{tab:instruction_pool}.

\begin{table*}[h]
\centering
\footnotesize
\resizebox{0.8\linewidth}{!}{%
\begin{tabular}{cp{0.28\linewidth}p{0.35\linewidth}p{0.28\linewidth}}
\toprule
\# & English & German & Portuguese \\
\midrule
1 & Determine the correctness of the student answer based on the information given.
  & Bestimmen Sie die Korrektheit der Schülerantwort basierend auf den gegebenen Informationen.
  & Determine a correção da resposta do aluno com base nas informações fornecidas. \\[3pt]
2 & Assess the response in relation to the given criteria.
  & Bewerten Sie die Antwort im Hinblick auf die angegebenen Kriterien.
  & Avalie a resposta em relação aos critérios fornecidos. \\[3pt]
3 & Determine if the answer aligns with the expected standards and provides a clear justification.
  & Bestimmen Sie, ob die Antwort den erwarteten Standards entspricht und eine klare Begründung liefert.
  & Determine se a resposta está alinhada com os padrões esperados e fornece uma justificativa clara. \\[3pt]
4 & Evaluate the answer based on the provided rubric and other context information.
  & Bewerten Sie die Antwort basierend auf der bereitgestellten Rubrik und anderen Kontextinformationen.
  & Avalie a resposta com base na rubrica fornecida e outras informações contextuais. \\[3pt]
5 & Score the response according to the rubric criteria and supporting materials.
  & Bewerten Sie die Antwort gemäß den Rubrikkriterien und unterstützenden Materialien.
  & Pontue a resposta de acordo com os critérios da rubrica e materiais de apoio. \\[3pt]
6 & Analyze whether the student's answer meets the requirements outlined in the rubric.
  & Analysieren Sie, welche Rubrikkriterien die Schülerantwort erfüllt.
  & Analise se a resposta do aluno atende aos requisitos descritos na rubrica. \\[3pt]
7 & Evaluate the completeness and correctness of the answer using the provided guidelines.
  & Bewerten Sie die Korrektheit der Antwort anhand der bereitgestellten Richtlinien.
  & Avalie a completude e a correção da resposta usando as diretrizes fornecidas. \\[3pt]
8 & Judge the quality of the response based on the rubric.
  & Beurteilen Sie die Qualität der Antwort basierend auf der Rubrik und den Referenzmaterialien.
  & Julgue a qualidade da resposta com base na rubrica. \\[3pt]
9 & Assess how well the student's answer addresses the key points in the rubric.
  & Bewerten Sie, wie gut die Schülerantwort die Frage beantwortet.
  & Avalie o quão bem a resposta do aluno aborda os pontos-chave da rubrica. \\[3pt]
10 & Rate the answer's alignment with the expected solution and grading criteria.
   & Bewerten Sie die Übereinstimmung der Antwort mit der erwarteten Lösung und den Bewertungskriterien.
   & Classifique o alinhamento da resposta com a solução esperada e os critérios de avaliação. \\[3pt]
\bottomrule
\end{tabular}}
\caption{Full instruction pool ($M=10$) in English, German, and Portuguese.}
\label{tab:instruction_pool}
\end{table*}

\begin{figure*}[t]
\centering
\footnotesize
\resizebox{0.8\linewidth}{!}{%
\begin{tabular}{@{}p{0.10\linewidth}p{0.28\linewidth}p{0.32\linewidth}p{0.26\linewidth}@{}}
\toprule
 & \textbf{English} & \textbf{German} & \textbf{Portuguese} \\
\midrule
\textbf{System} &
You are an expert grader of short student answers. You will receive a question, a sample solution, the student's answer, and a numbered list of rubric levels (starting at 0). Read everything carefully, briefly explain your reasoning, and then state which rubric level best matches the student's answer. &
Sie sind eine Expertin bzw.\ ein Experte f\"ur die Bewertung kurzer Sch\"ulerantworten. Sie erhalten eine Frage, eine Musterl\"osung, die Antwort der Sch\"ulerin bzw.\ des Sch\"ulers sowie eine nummerierte Liste von Rubrik-Stufen (beginnend bei 0). Lesen Sie alles sorgf\"altig, erl\"autern Sie kurz Ihre Begr\"undung und geben Sie anschlie{\ss}end an, welche Rubrik-Stufe am besten zur Sch\"ulerantwort passt. &
Voc\^{e} \'e uma especialista ou um especialista em avalia\c{c}\~{a}o de respostas curtas de alunos. Voc\^{e} receber\'a uma pergunta, uma solu\c{c}\~{a}o exemplar, a resposta do aluno e uma lista numerada de n\'iveis da rubrica (come\c{c}ando em 0). Leia tudo com aten\c{c}\~{a}o, explique brevemente o seu racioc\'inio e, em seguida, indique qual n\'ivel da rubrica melhor corresponde \`{a} resposta do aluno. \\[6pt]
\textbf{Output format} &
End your response with a final line in this exact format: \texttt{Final answer: <integer>}, where \texttt{<integer>} is the chosen rubric level (between 0 and \texttt{\{max\_score\}}, inclusive). &
Beenden Sie Ihre Antwort mit einer letzten Zeile in genau diesem Format: \texttt{Endg\"ultige Antwort: <Ganzzahl>}, wobei \texttt{<Ganzzahl>} die gew\"ahlte Rubrik-Stufe (zwischen 0 und \texttt{\{max\_score\}}, einschlie{\ss}lich) ist. &
Termine a sua resposta com uma \'ultima linha exatamente neste formato: \texttt{Resposta final: <inteiro>}, onde \texttt{<inteiro>} \'e o n\'ivel da rubrica escolhido (entre 0 e \texttt{\{max\_score\}}, inclusive). \\[6pt]
\midrule
\multicolumn{4}{@{}l}{\textbf{User template} (structure identical across languages; field labels follow \autoref{tab:field_translations}):} \\[2pt]
\multicolumn{4}{@{}l}{\texttt{\{question\_label\}: \{question\}}} \\
\multicolumn{4}{@{}l}{\texttt{\{solution\_label\}: \{sample\_solution\}}} \\
\multicolumn{4}{@{}l}{\texttt{\{answer\_label\}: \{answer\}}} \\[2pt]
\multicolumn{4}{@{}l}{Rubric levels / Rubrik-Stufen / N\'iveis da rubrica:} \\
\multicolumn{4}{@{}l}{\texttt{\{rubric\_lines\}}} \\[2pt]
\multicolumn{4}{@{}l}{\texttt{\{instruction\}}} \\
\bottomrule
\end{tabular}}
\caption{Zero-shot prompt templates in English, German, and Portuguese.}
\label{fig:zero_shot_prompts}
\end{figure*}

\FloatBarrier

\section{Zero-shot Prompt Templates} \label{ap:zero_shot_prompts}

\autoref{fig:zero_shot_prompts} shows the zero-shot prompt templates used for the LLM baseline in English, German, and Portuguese.

\FloatBarrier

\section{Hyperparameters and Model Choices} \label{ap:hparams_model_choices}
We report the full hyperparameter settings and model design choices used in our experiments in this section.

\subsection{Mono-Benchmark Training}

The backbone models used in our experiments are \textbf{Llama-3.2-1B-Instruct}\footnote{\url{https://huggingface.co/meta-llama/Llama-3.2-1B-Instruct}}, \textbf{Llama-3.2-3B-Instruct}\footnote{\url{https://huggingface.co/meta-llama/Llama-3.2-3B-Instruct}}, \textbf{Mistral-7B-v0.1}\footnote{\url{https://huggingface.co/mistralai/Mistral-7B-v0.1}}, and \textbf{Llama-3.1-8B-Instruct}\footnote{\url{https://huggingface.co/meta-llama/Llama-3.1-8B-Instruct}}.

\paragraph{\textsc{Ruspan}, Standard Classification, and Rubric Retrieval.}
All models are fine-tuned using LoRA with rank~64, $\alpha = 64$, dropout~0.1, and adapters applied to all linear modules.
We use a maximum sequence length of 2048 tokens and an effective batch size of 16.
Optimization uses a cosine learning-rate schedule with warmup over the first 1\% of training steps.
For decoder-only LLM backbones, we use paged AdamW in 32-bit; for BERT-style encoders, we use AdamW.
All experiments use a fixed random seed and are run on two NVIDIA A100 GPUs.
For \textsc{Alice-LP}, we train for 3 epochs, as we find this maximizes performance on unseen questions; all other models are trained for 5 epochs.
The learning rate is set to $2 \times 10^{-4}$ for Llama-3.2-1B-Instruct and Llama-3.2-3B-Instruct, and $5 \times 10^{-5}$ for Mistral-7B-v0.1 and Llama-3.1-8B-Instruct.
Checkpoints are selected by the highest macro-F1 on the development split.
For benchmarks without an official development split, we randomly sample a held-out development subset from the training split for checkpoint selection.
Note that Mistral-7B-v0.1 and Llama-3.1-8B-Instruct are not evaluated under the Rubric Retrieval setup, as the approach expands each instance into multiple answer-rubric pairs, making training and inference prohibitively expensive at this scale.

\paragraph{LLM Generation.}
Models are trained for 3 epochs with a learning rate of $1 \times 10^{-4}$ for Llama-3.2-1B-Instruct and Llama-3.2-3B-Instruct. All other hyperparameters follow the settings above.
At evaluation time, generated labels are decoded greedily with temperature~0.

\subsection{Joint Training and Cross-Benchmark Evaluation}
All fine-tuned models in the joint setting are trained for two epochs, which we find maximizes performance on unseen benchmarks. All other hyperparameters follow the mono-benchmark settings above.

\FloatBarrier

\FloatBarrier

\section{Ablation Studies} \label{ap:ablation}

\subsection{Rubric Quality and Use} \label{ap:ablation_rubric_quality}
To investigate rubric quality and rubric use, we conduct three sets of experiments.

First, we compare Sequence Classification with and without appended rubric text.
This tests whether simply exposing a fixed-head classifier to rubric descriptions is sufficient.

Second, for benchmarks without original rubric descriptions, we apply \textsc{Ruspan} but replace the detailed generated level 
descriptions with short label names (e.g., ``Incorrect'', ``Partially Correct'', ``Correct'').
This tests whether the model benefits from the semantic content of the generated rubric descriptions beyond the label names alone.
Note that \textsc{PT-ASAG} is excluded because this dataset only provides a numerical score, hence we cannot ground 
the generated rubrics to a concrete set of level names.

Third, for benchmarks with human-written rubrics 
(\textsc{ASAP-SAS}, \textsc{Alice-LP}), 
we use the same prompt to generate rubrics and compare them to the human-written ones.

\begin{table*}[h]
\centering
\scriptsize
\renewcommand{\arraystretch}{1.08}
\setlength{\tabcolsep}{3pt}
\resizebox{0.7\textwidth}{!}{%
\begin{tabular}{p{0.05\linewidth}p{0.40\linewidth}p{0.50\linewidth}}
\toprule
Level & Human-written rubric & LLM-generated rubric \\
\midrule
\multicolumn{3}{p{0.95\linewidth}}{\textbf{\textsc{ASAP-SAS}.} Question: After reading the group's procedure, describe what additional information you would need in order to replicate the experiment. Make sure to include at least three pieces of information.} \\
\midrule
0 & The response describes little or no accurate or relevant information from the acid rain investigation.
& Responses give no specific procedural details needed to replicate the acid-rain experiment and often include unrelated statements. They fail to identify any of the key missing items from the reference answers, such as type or amount of vinegar, sample identity or size, exposure time, container dimensions, rinsing procedure, or measurement methods. \\
1 & The response describes one additional piece of information that would be needed to accurately replicate the experiment.
& Responses identify exactly one specific procedural detail that is missing for replication, for example naming the type of vinegar, the exact volume of vinegar, or the duration the samples are submerged. They clearly state that single item but omit other essential parameters like sample size or shape, container specifications, rinsing steps, or measurement procedures. \\
2 & The response describes two additional pieces of information that would be needed to accurately replicate the experiment.
& Responses identify two distinct, specific pieces of missing procedural information, for example specifying both what the samples are and how much vinegar to use, or the exposure time and the instrument for measuring mass. They demonstrate partial sufficiency for replication but still omit at least one other critical element required for full reproducibility. \\
3 & The response describes three additional pieces of information that would be needed to accurately replicate the experiment.
& Responses list three specific procedural details that would enable replication, such as the material and size of the samples, the exact volume or concentration of vinegar to use, and the duration or environmental conditions for the exposure. They differ from lower scores by covering multiple different categories of missing information rather than repeating similar items. \\
\bottomrule
\end{tabular}}
\caption{Qualitative comparison of human-written and LLM-generated rubrics for an \textsc{ASAP-SAS} item.}
\label{tab:qual_rubric_asap}
\end{table*}

\begin{table*}[h]
\centering
\scriptsize
\renewcommand{\arraystretch}{1.08}
\setlength{\tabcolsep}{3pt}
\resizebox{0.7\textwidth}{!}{%
\begin{tabular}{p{0.05\linewidth}p{0.40\linewidth}p{0.50\linewidth}}
\toprule
Level & Human-written rubric & LLM-generated rubric \\
\midrule
\multicolumn{3}{p{0.95\linewidth}}{\textbf{\textsc{Alice-LP}.} Question: \textcolor{gray}{Welche wissenschaftlichen Fragen fallen dir zur Färbung und Musterung der Guppys ein? Tipp: Wissenschaftliche Fragen sollten überprüfbar sein, beispielsweise durch Experimente.} \textit{English: What scientific questions occur to you about the coloration and patterning of guppies? Tip: Scientific questions should be testable, for example through experiments.}} \\
\midrule
0 & \textcolor{gray}{Die Schüler:innen formulieren keine wissenschaftliche Frage zur Färbung und Musterung der Guppys genannt.}
(\textit{\textbf{English}: Students do not formulate a scientific question about the coloration and patterning of the guppies.})
& \textcolor{gray}{Antworten zeigen ein grundlegendes Verständnis für die Färbung und Musterung der Guppys, jedoch fehlen klare, überprüfbare wissenschaftliche Fragen. Die Formulierungen sind ungenau und betreffen oft allgemeine Beobachtungen oder Vermutungen ohne spezifische wissenschaftliche Ansätze.}
(\textit{\textbf{English}: Responses show a basic understanding of the coloration and patterning of guppies, but clear, testable scientific questions are missing. The formulations are imprecise and often concern general observations or conjectures without specific scientific approaches.}) \\
1 & \textcolor{gray}{Die Schüler:innen formulieren eine wissenschaftliche Frage zur Färbung und Musterung der Guppys genannt mit Bezug auf Variation zwischen Individuen oder Geschlechtern.}
(\textit{\textbf{English}: Students formulate one scientific question about the coloration and patterning of the guppies with reference to variation between individuals or sexes.})
& \textcolor{gray}{Antworten enthalten einige überprüfbare Fragen zur Färbung und Musterung der Guppys, jedoch sind diese oft weniger präzise und weniger tiefgehend. Die Fragen zeigen ein gewisses Verständnis für Unterschiede zwischen Männchen und Weibchen, thematisieren jedoch nicht alle relevanten wissenschaftlichen Aspekte oder Zusammenhänge.}
(\textit{\textbf{English}: Responses contain some testable questions about the coloration and patterning of guppies, but these are often less precise and less in-depth. The questions show some understanding of differences between males and females, but do not address all relevant scientific aspects or relationships.}) \\
2 & \textcolor{gray}{Die Schüler:innen formulieren mindestens zwei wissenschaftliche Fragen zur Färbung und Musterung der Guppys genannt mit Bezug auf Variation zwischen Individuen und Geschlechtern.}
(\textit{\textbf{English}: Students formulate at least two scientific questions about the coloration and patterning of the guppies with reference to variation between individuals and sexes.})
& \textcolor{gray}{Antworten formulieren mehrere präzise und überprüfbare wissenschaftliche Fragen zur Färbung und Musterung der Guppys. Die Fragen zeigen ein tiefes Verständnis für biologische Konzepte wie Geschlechtsdimorphismus, genetische Variationen und deren Auswirkungen auf das Überleben und die Fortpflanzung der Guppys.}
(\textit{\textbf{English}: Responses formulate several precise and testable scientific questions about the coloration and patterning of guppies. The questions show a deep understanding of biological concepts such as sexual dimorphism, genetic variation, and their effects on guppy survival and reproduction.}) \\
\bottomrule
\end{tabular}}
\caption{Qualitative comparison of human-written and LLM-generated rubrics for an \textsc{Alice-LP} item. The original German text is shown in gray and followed by an English translation.}
\label{tab:qual_rubric_alice}
\end{table*}

The examples in \autoref{tab:qual_rubric_asap} and \autoref{tab:qual_rubric_alice} qualitatively compare human-written and LLM-generated rubric versions.
The generated rubrics add concrete examples, explicit contrasts between adjacent levels, and response-oriented phrasing.
However, they also introduce more interpretive content than the original rubrics, which can make the criteria more detailed but may also shift emphasis away from the concise distinctions used by human annotators.
This is highly salient in the example of \textsc{Alice-LP}, as the human-written rubrics 
focus on the number of scientific questions and the presence of variation, while
the generated rubrics emphasize the depth of understanding and specific biological concepts mentioned in the responses. Thus, 
without understanding the original pedagogical intent and annotation guidelines, 
the generated rubrics may not align with the human-written ones in terms of what aspects of the responses are most important for distinguishing between levels.

\begin{table*}[h]
\centering
\small
\renewcommand{\arraystretch}{0.95}
\setlength{\tabcolsep}{3pt}
\resizebox{0.7\textwidth}{!}{%
\begin{tabular}{ll cccc cc}
\toprule
\multirow{2}{*}{Model} & \multirow{2}{*}{Variant}
    & \multicolumn{1}{c}{iStudio}
    & \multicolumn{2}{c}{SciEntsBank}
    & \multicolumn{1}{c}{Beetle}
    & \multicolumn{1}{c}{ASAP-SAS}
    & \multicolumn{1}{c}{Alice-LP} \\
\cmidrule(lr){3-3}\cmidrule(lr){4-5}\cmidrule(lr){6-6}\cmidrule(lr){7-7}\cmidrule(lr){8-8}
 & & UA & UQ & UD & UQ & UA & UQ \\
\midrule
\multirow{5}{*}{\textit{Llama-3.2-1B-Instruct}} & \textsc{concat}
    & 84.8 & \cellcolor{red!15}65.3 & 64.5 & \multicolumn{1}{c:}{64.5}
    & 77.8 & \cellcolor{red!15}61.3 \\
    & \textsc{diff}
    & \cellcolor{red!15}87.2 & \cellcolor{red!15}30.3 & \cellcolor{red!15}26.9 & \multicolumn{1}{c:}{\cellcolor{red!15}65.6}
    & 78.5 & 65.6 \\
    & \textsc{condiff}
    & \cellcolor{red!15}86.8 & 66.9 & 65.9 & \multicolumn{1}{c:}{\cellcolor{red!15}66.2}
    & \cellcolor{red!15}76.9 & \cellcolor{red!15}62.6 \\
    & \textsc{no-align}
    & \cellcolor{red!15}87.2 & 64.3 & \cellcolor{red!15}63.2 & \multicolumn{1}{c:}{65.5}
    & 76.9 & \cellcolor{red!15}62.1 \\
\hdashline
    & Seq.\ Class. w/o Rub.
    & 85.4 & \cellcolor{red!15}66.2 & \cellcolor{red!15}64.9 & \cellcolor{red!15}67.0
    & N/A & \cellcolor{red!15}63.1 \\
\midrule
\multirow{5}{*}{\textit{Llama-3.2-3B-Instruct}} & \textsc{concat}
    & 87.1 & 67.7 & 67.5 & \multicolumn{1}{c:}{67.8}
    & \cellcolor{red!15}80.0 & \cellcolor{red!15}64.1 \\
    & \textsc{diff}
    & \cellcolor{red!15}88.4 & 63.8 & \cellcolor{red!15}67.1 & \multicolumn{1}{c:}{66.3}
    & 79.8 & \cellcolor{red!15}61.7 \\
    & \textsc{condiff}
    & \cellcolor{red!15}87.0 & 68.4 & 68.4 & \multicolumn{1}{c:}{66.4}
    & \cellcolor{red!15} 78.2 & \cellcolor{red!15}66.0 \\
    & \textsc{no-align}
    & 84.1 & 66.7 & 67.2 & \multicolumn{1}{c:}{\cellcolor{red!15}66.8}
    & 79.2 & \cellcolor{red!15}62.5 \\
\hdashline
    & Seq.\ Class. w/o Rub.
    & \cellcolor{red!15}88.3 & \cellcolor{red!15}67.5 & \cellcolor{red!15}66.6 & \cellcolor{red!15}67.6
    & N/A & 64.2 \\
\bottomrule
\end{tabular}}
\caption{Quality-control ablations for rubric descriptions (macro-F1). For \textsc{Ruspan} variants, the left block reports performance when detailed generated rubric descriptions are replaced with short label names, and the right block reports performance when human-written rubric descriptions are replaced with LLM-generated rubric descriptions. For Sequence Classification, the row reports performance without appended rubric text, with N/A for \textsc{ASAP-SAS}. Light red marks cases where the ablated condition is worse for human-written rubric replacement or better for label-name and no-rubric ablations.}
\label{tab:rubric_quality_ablation}
\end{table*}

For Sequence Classification, \autoref{tab:rubric_quality_ablation} shows that appending rubric text alone is not sufficient:
removing rubrics often leaves performance unchanged or even improved,
such as on \textsc{SciEntsBank} with the 1B backbone and on \textsc{iStudio} with the 3B backbone.
Thus, rubric text is useful when the model explicitly binds it to candidate score levels, rather than simply appending it to a fixed-head classifier input.
Although rubric conditioning does not always yield the best result under mono-benchmark training, transfer to unseen questions or to benchmarks with different rubric structures is a more realistic educational setting, where scoring systems must generalize beyond the questions and rubrics observed during training.

For benchmarks without original rubric descriptions, detailed LLM-generated rubrics do not always outperform short label names, yet a pattern emerges. 
Across alignment functions, the \textsc{diff} variant underperforms when detailed rubrics are used, indicating limited 
semantic representation capacity of the generated rubrics. In terms of models, Llama-3.2-3B-Instruct clearly benefits more from
detailed rubric descriptions, suggesting that larger models may be better at leveraging the semantic content of the generated rubrics.
As for the differences between datasets, \textsc{iStudio}-UA seems to benefit less from detailed LLM-generated rubrics. We hypothesize
that this is because it is tested on questions seen during training, making it essentially an easier setting where the model can rely more on pattern recognition and less on semantic understanding of the rubrics. 

The results show that replacing human-written rubrics with LLM-generated rubrics does not consistently preserve performance.
On \textsc{ASAP-SAS}, the effect is mixed: generated rubrics slightly improve several settings, but they also underperform for \textsc{condiff} with Llama-3.2-1B-Instruct and \textsc{concat} with Llama-3.2-3B-Instruct.
The degradation is clearer on the \textsc{Alice-LP} UQ split, where generated rubrics underperform human-written rubrics in most settings.
This suggests a limitation of automatic rubric generation from information only available in the public datasets: without more detailed information about the target task, scoring conventions, and question-specific expectations, LLM-generated rubrics may omit distinctions that are important for reliable scoring.

\subsection{Sample-Solution Ablation} \label{ap:ablation_sample_solution}

We examine whether the LLM-generated sample solution contributes to performance by removing it from the model input and reporting the change in macro-F1 relative to the full-input setting.
This ablation applies only to \textsc{ASAP-SAS}, the only benchmark for which we generate sample solutions with an LLM.

\begin{table}[H]
\centering
\small
\setlength{\tabcolsep}{5pt}
\begin{tabular}{llr}
\toprule
Model & Variant & $\Delta$F1 \\
\midrule
\multirow{2}{*}{\textit{Llama-3.2-1B-Instruct}}
    & \textsc{Ruspan}-\textsc{concat}  & $-1.1$ \\
    & \textsc{Ruspan}-\textsc{condiff} & $-0.1$ \\
\midrule
\multirow{2}{*}{\textit{Llama-3.2-3B-Instruct}}
    & \textsc{Ruspan}-\textsc{concat}  & $+0.1$ \\
    & \textsc{Ruspan}-\textsc{condiff} & $+0.7$ \\
\midrule
\multicolumn{2}{l}{Mean} & $-0.1$ \\
\bottomrule
\end{tabular}
\caption{Effect of removing the LLM-generated sample solution from \textsc{ASAP-SAS} inputs. Values are changes in macro-F1 relative to the full-input setting (ablated minus full). This ablation applies only to \textsc{ASAP-SAS}.}
\label{tab:sample_solution_ablation}
\end{table}

Removing the generated sample solution changes F1 by only $-0.1$ on average, with variation across backbones and alignment functions.
We retain it to keep the \textsc{ASAP-SAS} input format consistent with the other ASAS benchmarks.

\FloatBarrier

\section{Label Distributions} \label{ap:label_distributions}

\begin{figure*}[t]
\centering
\includegraphics[width=\textwidth]{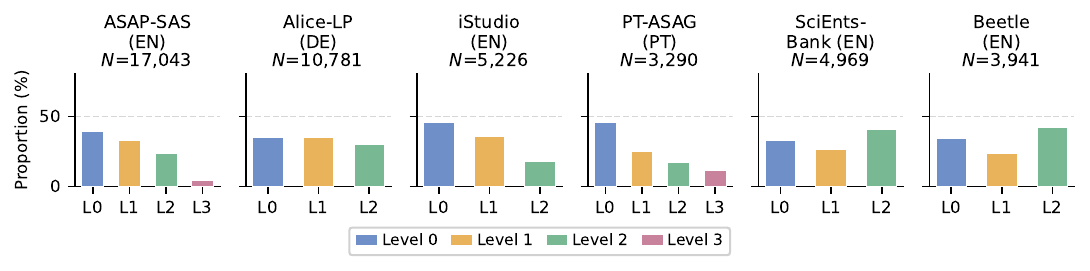}
\caption{Label distributions for the six ASAS benchmarks.}
\label{fig:label_distribution}
\end{figure*}

\section{Non-ASAS Datasets} \label{ap:non_asas}

\label{app:non_asas_datasets}

We evaluate on seven non-ASAS benchmarks covering commonsense reasoning,
figurative-language interpretation, stance detection, 
sentiment analysis, news topic classification, and scientific edit-intent classification, 
ranging from relatively simple tasks that may not require complex label descriptions and rely mostly on lexical cues in the input texts (e.g., sentiment analysis and
topic classification) to more complex tasks that may benefit from detailed rubric descriptions (e.g., commonsense reasoning and stance
detection).

We compare sequence classification baselines (with and without rubric text) and four \textsc{Ruspan} variants (\textsc{Ruspan}-\textsc{condiff}, \textsc{Ruspan}-\textsc{concat}, \textsc{Ruspan}-\textsc{diff}, and \textsc{Ruspan}-\textsc{no-align}), with the exception of PIQA and FigQA, where the candidate solutions serve as the rubric text and cannot be removed without changing the task.
\autoref{tab:non_asas_results} reports results for the seven non-ASAS benchmarks.

\paragraph{PIQA.}
A physical commonsense reasoning benchmark~\cite{bisk2020piqa}. Given a goal, the model must select the more plausible of two candidate solutions. The task is cast as multiple choice.

\paragraph{xStance.}
A multilingual political stance detection dataset~\cite{vamvas2020xstance} covering German, French, and Italian. Given a policy question and a short comment, the model decides whether the author \textbf{supports} or \textbf{opposes} the claim. 

The training set is in German and French, while the test set is Italian; this setup tests the model's ability to generalize across languages.

\paragraph{SemEval-2016 Task 6.}
A tweet stance detection benchmark~\cite{mohammad2016semeval}. Given a tweet and a named target, the model classifies the stance as \textbf{Favor}, \textbf{Against}, or \textbf{None}. The test split covers both seen targets (shared with training) and an unseen target (\textit{Donald Trump}), allowing evaluation under both in-domain and zero-shot target conditions.

\paragraph{FigQA.}
A figurative-language multiple-choice benchmark~\cite{liu2022figqa}. Given a figurative sentence, the model selects the ending that best matches its intended meaning.

\paragraph{IMDB.}
A binary movie-review sentiment benchmark~\cite{maas2011imdb}. The model classifies a review as expressing \textbf{negative} or \textbf{positive} sentiment. Due to the large size of this dataset, we randomly sample 30\% of the training split for all experiments; our focus is on comparing models rather than maximizing absolute performance.

\paragraph{AG News.}
A four-way news topic classification benchmark~\cite{zhang2015agnews}. 

Given a short news article, 
the model classifies it as \textbf{World}, \textbf{Sports}, \textbf{Business}, or \textbf{Sci/Tech}.
We similarly sample 30\% of the training split due to the large dataset size.

\paragraph{EIC.}
The Edit Intent Classification benchmark~\cite{ruan2024largelanguagemodelsgoodclassifier} tests whether the rubric-label interface extends to a specialized discourse task. Given an original and a revised scientific text, the model identifies the purpose of the edit as \textbf{Claim}, \textbf{Clarity}, \textbf{Fact/Evidence}, \textbf{Grammar}, or \textbf{Other}.

\begin{figure*}[t]
\centering
\footnotesize

\begin{minipage}[t]{0.47\linewidth}
\noindent\fbox{\parbox{0.95\linewidth}{%
  \textbf{PIQA}\\[2pt]
  \texttt{<goal>} \textit{When boiling butter, when it's ready, you can} \texttt{</goal>}\\[3pt]
  \colorbox{cyan!20}{\parbox[t]{\dimexpr\linewidth-6pt\relax}{\ttfamily <rubric> Pour it onto a plate </rubric>}}\\[1pt]
  \colorbox{cyan!20}{\parbox[t]{\dimexpr\linewidth-6pt\relax}{\ttfamily <rubric> Pour it into a jar </rubric>}}\\[3pt]
  \textit{Select the most plausible solution\ldots}
}}
\end{minipage}
\hfill
\begin{minipage}[t]{0.47\linewidth}
\noindent\fbox{\parbox{0.95\linewidth}{%
  \textbf{xStance}\\[2pt]
  \texttt{<question>} \textit{Eine Volksinitiative fordert die Begrenzung der Bauzonen \ldots\ Befürworten Sie?} \texttt{</question>}\\
  \texttt{<comment>} \textit{Eine fixe Grösse verbieten, ist das falsche Mittel} \texttt{</comment>}\\[3pt]
  \colorbox{cyan!20}{\parbox[t]{\dimexpr\linewidth-6pt\relax}{\ttfamily <rubric> the author is supportive of the claim </rubric>}}\\[1pt]
  \colorbox{cyan!20}{\parbox[t]{\dimexpr\linewidth-6pt\relax}{\ttfamily <rubric> the author is against the claim </rubric>}}\\[3pt]
  \textit{Determine whether the comment supports\ldots}
}}
\end{minipage}

\vspace{4pt}

\begin{minipage}[t]{0.47\linewidth}
\noindent\fbox{\parbox{0.95\linewidth}{%
  \textbf{SemEval-2016 Task 6}\\[2pt]
  \texttt{<target>} \textit{Atheism} \texttt{</target>}\\
  \texttt{<tweet>} \textit{dear lord thank u for all of ur blessings \ldots\ \#SemST} \texttt{</tweet>}\\[3pt]
  \colorbox{cyan!20}{\parbox[t]{\dimexpr\linewidth-6pt\relax}{\ttfamily <rubric> the author is supportive of Atheism </rubric>}}\\[1pt]
  \colorbox{cyan!20}{\parbox[t]{\dimexpr\linewidth-6pt\relax}{\ttfamily <rubric> the author is against Atheism </rubric>}}\\[1pt]
  \colorbox{cyan!20}{\parbox[t]{\dimexpr\linewidth-6pt\relax}{\ttfamily <rubric> the author is neutral to Atheism </rubric>}}\\[3pt]
  \textit{Determine whether the tweet supports\ldots}
}}
\end{minipage}
\hfill
\begin{minipage}[t]{0.47\linewidth}
\noindent\fbox{\parbox{0.95\linewidth}{%
  \textbf{FigQA}\\[2pt]
  \texttt{<sentence>} \textit{Her word had the strength of titanium.} \texttt{</sentence>}\\[3pt]
  \colorbox{cyan!20}{\parbox[t]{\dimexpr\linewidth-6pt\relax}{\ttfamily <rubric> Her promises can be believed. </rubric>}}\\[1pt]
  \colorbox{cyan!20}{\parbox[t]{\dimexpr\linewidth-6pt\relax}{\ttfamily <rubric> Her promises cannot be trusted. </rubric>}}\\[3pt]
  \textit{Choose the ending that best matches\ldots}
}}
\end{minipage}

\vspace{4pt}

\begin{minipage}[t]{0.47\linewidth}
\noindent\fbox{\parbox{0.95\linewidth}{%
  \textbf{IMDB}\\[2pt]
  \texttt{<review>} \textit{I rented I AM CURIOUS-YELLOW \ldots\ this film does not have much of a plot.} \texttt{</review>}\\[3pt]
  \colorbox{cyan!20}{\parbox[t]{\dimexpr\linewidth-6pt\relax}{\ttfamily <rubric> the review expresses negative sentiment </rubric>}}\\[1pt]
  \colorbox{cyan!20}{\parbox[t]{\dimexpr\linewidth-6pt\relax}{\ttfamily <rubric> the review expresses positive sentiment </rubric>}}\\[3pt]
  \textit{Classify the sentiment of the review\ldots}
}}
\end{minipage}
\hfill
\begin{minipage}[t]{0.47\linewidth}
\noindent\fbox{\parbox{0.95\linewidth}{%
  \textbf{AG News}\\[2pt]
  \texttt{<text>} \textit{Shares of Google rose 5\% after strong quarterly earnings driven by ad revenue.} \texttt{</text>}\\[3pt]
  \colorbox{cyan!20}{\parbox[t]{\dimexpr\linewidth-6pt\relax}{\ttfamily <rubric> the article is about world news, international affairs, politics, or global events </rubric>}}\\[1pt]
  \colorbox{cyan!20}{\parbox[t]{\dimexpr\linewidth-6pt\relax}{\ttfamily <rubric> the article is about sports, games, athletes, teams, or sporting events </rubric>}}\\[1pt]
  \colorbox{cyan!20}{\parbox[t]{\dimexpr\linewidth-6pt\relax}{\ttfamily <rubric> the article is about business, finance, markets, companies, or the economy </rubric>}}\\[1pt]
  \colorbox{cyan!20}{\parbox[t]{\dimexpr\linewidth-6pt\relax}{\ttfamily <rubric> the article is about science, technology, computing, research, or innovation </rubric>}}\\[3pt]
  \textit{Classify the news article into one of\ldots}
}}
\end{minipage}

\vspace{4pt}

\begin{minipage}[t]{\linewidth}
\noindent\fbox{\parbox{0.97\linewidth}{%
  \textbf{EIC}\\[2pt]
  \texttt{<old>} \textit{For MBTI, users were able to provide multiple texts, we report unique users in parentheses.} \texttt{</old>}\\
  \texttt{<new>} \textit{[empty]} \texttt{</new>}\\[3pt]
  \colorbox{cyan!20}{\parbox[t]{\dimexpr\linewidth-6pt\relax}{\ttfamily <rubric> the edit modifies or introduces a claim or argument </rubric>}}\\[1pt]
  \colorbox{cyan!20}{\parbox[t]{\dimexpr\linewidth-6pt\relax}{\ttfamily <rubric> the edit improves the clarity or readability of the text </rubric>}}\\[1pt]
  \colorbox{cyan!20}{\parbox[t]{\dimexpr\linewidth-6pt\relax}{\ttfamily <rubric> the edit adds, removes, or modifies factual information or supporting evidence </rubric>}}\\[1pt]
  \colorbox{cyan!20}{\parbox[t]{\dimexpr\linewidth-6pt\relax}{\ttfamily <rubric> the edit corrects grammatical, spelling, or punctuation errors </rubric>}}\\[1pt]
  \colorbox{cyan!20}{\parbox[t]{\dimexpr\linewidth-6pt\relax}{\ttfamily <rubric> the edit serves another purpose not covered by the above categories </rubric>}}\\[3pt]
  \textit{Classify the edit intent for the source and target text\ldots}
}}
\end{minipage}
\caption{Truncated input examples for the seven non-ASAS benchmarks. Rubric (label) descriptions are highlighted in \colorbox{cyan!20}{blue}.}

\label{fig:non_asas_input_examples}
\end{figure*}

\begin{table*}[!t]
\centering
\small
\setlength{\tabcolsep}{3pt}
\renewcommand{\arraystretch}{0.95}
\resizebox{0.7\textwidth}{!}{%
\begin{tabular}{ll c c cc c c c c}
\toprule
\multirow{2}{*}{Model} & \multirow{2}{*}{Method}
    & \multicolumn{1}{c}{\textsc{PIQA}}
    & \multicolumn{1}{c}{\textsc{xStance}}
    & \multicolumn{2}{c}{\textsc{SemEval-2016 Task~6}}
    & \multicolumn{1}{c}{\textsc{FigQA}}
    & \multicolumn{1}{c}{\textsc{IMDB}}
    & \multicolumn{1}{c}{\textsc{AG News}}
    & \multicolumn{1}{c}{\textsc{EIC}} \\
\cmidrule(lr){3-3}\cmidrule(lr){4-4}\cmidrule(lr){5-6}\cmidrule(lr){7-7}\cmidrule(lr){8-8}\cmidrule(lr){9-9}\cmidrule(lr){10-10}
 & & & \textit{Italian} & \textit{Seen Targets} & \textit{Donald Trump} & & & & \\
\midrule
\multirow{6}{*}{\textit{Llama-3.2-1B-Instruct}}
    & Seq.\ Class.\ w/o Rub. & N/A & 78.4 & 73.6 & \textbf{61.3} & N/A & \textbf{96.4} & \textbf{95.2} & 80.2 \\
    & Seq.\ Class.\ w/ Rub.  & 78.3 & 78.8 & 73.7 & 45.8 & 92.3 & 95.8 & 95.1 & \textbf{81.5} \\
\cdashline{2-10}
    & \textsc{Ruspan}-\textsc{concat}   & \textbf{81.7} & \textbf{80.6} & 74.0 & 43.1 & \textbf{92.8} & 95.8 & 94.8 & 80.8 \\
    & \textsc{Ruspan}-\textsc{diff}     & 80.8 & 80.4 & 73.7 & 53.9 & 92.6 & 96.0 & 95.0 & 80.7 \\
    & \textsc{Ruspan}-\textsc{condiff}  & 80.5 & 79.7 & \textbf{76.1} & 60.5 & \textbf{92.8} & 96.3 & 95.1 & 80.5 \\
    & \textsc{Ruspan}-\textsc{no-align} & 81.6 & 80.5 & 75.9 & 42.3 & 92.4 & 95.9 & 95.0 & 80.5 \\
\midrule
\multirow{6}{*}{\textit{Llama-3.2-3B-Instruct}}
    & Seq.\ Class.\ w/o Rub. & N/A & 82.9 & 76.3 & 42.2 & N/A & \textbf{96.8} & 94.7 & \textbf{83.7} \\
    & Seq.\ Class.\ w/ Rub.  & 85.4 & 84.7 & 77.4 & 49.6 & 94.6 & \textbf{96.8} & \textbf{94.8} & 82.4 \\
\cdashline{2-10}
    & \textsc{Ruspan}-\textsc{concat}   & 87.8 & \textbf{84.8} & 76.8 & 64.1 & 95.5 & 96.4 & 94.6 & 83.5 \\
    & \textsc{Ruspan}-\textsc{diff}     & \textbf{88.3} & 84.7 & 78.6 & 53.6 & 95.4 & 96.7 & 94.5 & 83.0 \\
    & \textsc{Ruspan}-\textsc{condiff}  & 86.3 & 84.6 & 78.5 & \textbf{70.2} & 95.9 & \textbf{96.8} & 94.6 & 82.8 \\
    & \textsc{Ruspan}-\textsc{no-align} & 85.8 & \textbf{84.8} & \textbf{79.5} & 69.6 & \textbf{96.1} & 96.7 & 94.7 & 83.4 \\
\bottomrule
\end{tabular}}
\caption{Available non-ASAS benchmark results (macro-F1). \textbf{Bold} indicates the best completed result within each model group.}
\label{tab:non_asas_results}
\end{table*}

\autoref{tab:non_asas_results} yields three observations.

\paragraph{Saturation tasks are unaffected by rubric representations.}
On \textsc{IMDB} and \textsc{AG News}, all methods achieve near-identical macro-F1 (94 to 97\%), regardless of whether rubric text is present or how it is encoded.
These tasks have short, unambiguous label names whose semantics are fully recoverable from training data alone, leaving no room for detailed rubric descriptions to contribute.

\paragraph{Rubric text benefits classification only when coupled with per-label encoding.}
The \textsc{SemEval-2016 Task~6} Donald Trump split, which tests an unseen target at evaluation time, is the most informative condition.
For the 3B backbone, Seq.\ Class.\ w/o Rub.\ scores 42.2, while \textsc{Ruspan}-\textsc{condiff} and \textsc{Ruspan}-\textsc{no-align} reach 70.2 and 69.6.
Yet Seq.\ Class.\ w/ Rub.\ scores only 49.6, confirming that appending rubric text to a fixed-head classifier does not transfer the semantic content of label descriptions to the prediction.
The gain is specific to the per-label span encoding of \textsc{Ruspan}, which forces the model to bind answer representations to individual rubric spans rather than treating the rubric as undifferentiated context.
For the 1B backbone the effect is weaker: Seq.\ Class.\ w/o Rub.\ already reaches 61.3 by exploiting training-set lexical patterns, and rubric-based gains are less consistent, suggesting that the smaller model has limited capacity for compositional reasoning over unseen label descriptions.

\paragraph{\textsc{Ruspan} does not hurt on tasks where rubrics add no value.}
Across tasks where Seq.\ Class.\ w/o Rub.\ is competitive (EIC, IMDB, AG News), \textsc{Ruspan} variants match or come within 1 point of the baseline, indicating that the span-encoding overhead does not introduce a systematic penalty when rubric descriptions carry no additional discriminative signal.

\FloatBarrier
 
\end{document}